\documentclass{article} 
\usepackage{arxiv,times}

\usepackage{amsmath,amsfonts,bm}

\def\eqref#1{equation~\ref{#1}}

\def\1{\bm{1}}

\DeclareMathAlphabet{\mathsfit}{\encodingdefault}{\sfdefault}{m}{sl}
\SetMathAlphabet{\mathsfit}{bold}{\encodingdefault}{\sfdefault}{bx}{n}

\usepackage{hyperref}
\usepackage{url}
\usepackage{booktabs}
\usepackage{graphicx}
\usepackage{pifont}          
\usepackage{xcolor}
\usepackage{multirow}

\title{SonoWorld: Towards Interactive Pose-Controllable Ultrasound World Models}

\author{Zefan Yang$^{1,2}$\thanks{This work was carried out during the internship of the author at United Imaging Intelligence, Boston, MA.} , Zheyuan Zhang$^{1}$\thanks{Corresponding author.} , Boris Maihe$^{1}$, Yikang Liu$^{1}$, Hang Yu$^{1}$, Ankush Mukherjee$^{1}$, \\
\textbf{Xiao Chen$^{1}$, Pingkun Yan$^{2}$, Terrence Chen$^{1}$, Shanhui Sun$^{1}$} \\
$^{1}$United Imaging Intelligence, Boston, MA, USA \\
$^{2}$Rensselaer Polytechnic Institute, Troy, NY, USA \\
{\small\texttt{\{zefan.yang, zheyuan.zhang01, boris.maihe, yikang.liu, hang.yu02\}@uii-ai.com}} \\
{\small\texttt{\{ankush.mukherjee, xiao.chen01, terrence.chen\}@uii-ai.com}}, {\small\texttt{yanp2@rpi.edu}} \\
{\small\texttt{shanhui.sun@uii-ai.com}} \\
}

\begin{document}

\maketitle
\begin{figure}[ht]
    \centering
    \includegraphics[width=0.95\linewidth]{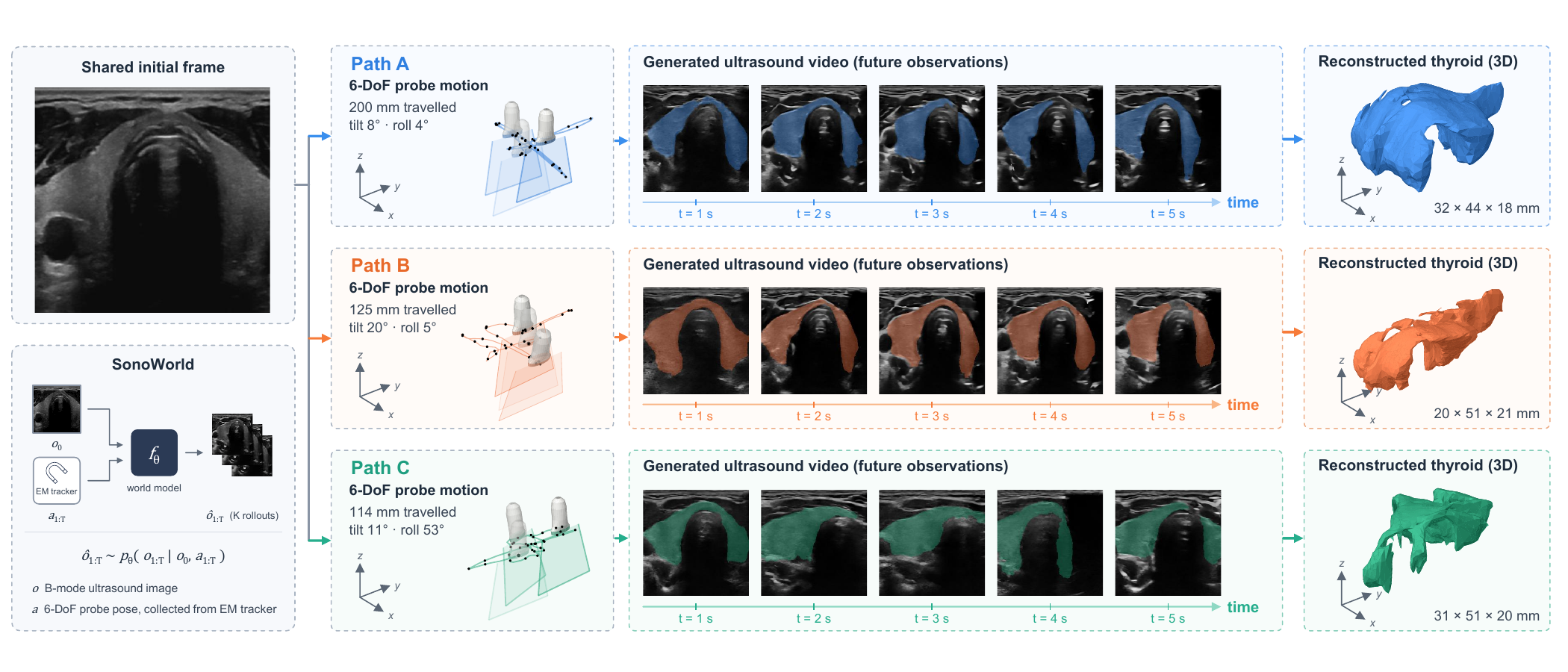}
    \caption{Starting from the same initial ultrasound frame,  SonoWorld generates distinct sequences of future ultrasound views along different 6-DoF probe scan paths, which captures how probe motion determines the evolving image appearance and ultimately the reconstructed 3D thyroid anatomy.}
    \label{fig:placeholder}
\end{figure}

\begin{abstract}

Robotic ultrasound can standardize image acquisition and reduce operator workload and variability, but policy learning remains limited by the lack of realistic, controllable simulation environments: physical phantoms and acoustic simulators exhibit a substantial domain gap from clinical images, and trial-and-error exploration on patients is unsafe. While controllable world models have advanced rapidly in natural-image domains, ultrasound world modeling remains challenging because each frame captures only a cross-section of unobserved 3D anatomy, causing small probe motions to induce large appearance changes. Existing approaches rely on robot-tracked poses and latent-space prediction, but do not directly generate the observations required for interactive control. We propose SonoWorld, a causal ultrasound world model that learns video dynamics from unlabeled clinical videos and grounds them to 6-DoF control using a small set of freehand clinical scans tracked with an electromagnetic (EM) sensor. SonoWorld is trained in three stages: pretraining a latent-action-conditioned video diffusion model on around 6.9 million unlabeled multi-organ ultrasound frames; grounding the learned video dynamics to metric probe poses with a small EM-tracked thyroid dataset; and applying distribution matching distillation to enable causal, long-horizon autoregressive generation with selectively caching eviction tokens at inference time. We evaluate controllability through pose-following fidelity, round-trip consistency, and axis-specific control under synthetic actions on held-out tracked scans, and further validate it with a user study. In model-based action discrimination, SonoWorld correctly ranks the target-directed probe motion over an equal-magnitude opposite motion in up to 74.8\% of trials. To our knowledge, SonoWorld is the first causal, pose-controllable ultrasound world model learned primarily from freehand clinical videos, providing a data-driven interactive environment for robotic ultrasound data acquisition and policy evaluation. 

\end{abstract}

\section{Introduction}

Ultrasound is among the most widely used diagnostic imaging modalities, valued for its real-time imaging, absence of ionizing radiation, portability, and low cost~\citep{wang2020anatomy,jiang2023robotic}. Unlike many imaging modalities, however, ultrasound acquisition is inherently interactive: the observed image depends on the six-degree-of-freedom (6-DoF) pose of a hand-held probe, which the operator continuously adjusts to navigate anatomy and acquire diagnostically standard views. Consequently, image quality and diagnostic yield depend heavily on operator expertise, contributing to substantial inter-operator variability and a persistent shortage of trained sonographers~\citep{jiang2023robotic}. Robotic ultrasound offers a promising path toward standardized image acquisition, reduced operator workload, and broader access to expert-level scanning, particularly in remote and resource-limited settings~\citep{du2024robot}. Realizing this potential requires robotic scanning policies that map observed ultrasound images to probe motions, whose development and evaluation in turn require realistic interactive environments in which the visual consequences of probe actions can be observed.

Existing environments for robotic ultrasound expose a fundamental trade-off between controllability, realism, and safety. 
Physical phantoms and physics-based simulators permit controlled exploration over probe poses, but their synthetic anatomy and image formation leave a substantial domain gap from clinical ultrasound~\citep{shams2008realtime, burger2013realtime, gao2009fast}. Re-slicing reconstructed 3D ultrasound volumes provides more realistic observations, but requires subject-specific tracked acquisitions and cannot faithfully reproduce view-dependent appearance changes under arbitrary probe orientations~\citep{lasso2014plus}. Scanning real patients provides observations from the true clinical distribution, yet repeated trial-and-error exploration is neither practical nor safe. An alternative line of work bypasses explicit simulation and learns probe guidance or autonomous scanning directly from tracked clinical scans and expert demonstrations~\citep{jiang2024cardiaccopilot,jiang2024structure,yue2025echoworld,su2024fully,jiang2025towards}. While effective, this paradigm depends on collecting real image--action trajectories at scale and does not provide a reusable environment for evaluating counterfactual probe actions before they are executed on patients. These limitations motivate a controllable generative environment learned directly from clinical ultrasound videos, capable of synthesizing realistic future observations in response to probe actions without tracking each patient.

Controllable world models offer a promising data-driven alternative by learning to predict future observations conditioned on actions, and have emerged as powerful interactive simulators for games, autonomous driving, and robotic manipulation~\citep{hou2026world,ding2025understanding,gao2026dreamdojo}. Ultrasound, however, presents a distinct challenge: each frame captures only a 2D cross-section of unobserved 3D anatomy, such that small probe motions, particularly out-of-plane translations and rotations, can induce large changes in image appearance. Existing ultrasound world models address this challenge using robot-tracked acquisitions~\citep{fan2026action} and latent-space prediction~\citep{yue2025echoworld, jiang2024cardiaccopilot}, but do not directly generate future ultrasound observations from freehand 6-DoF probe actions. Consequently, a causal, pose-controllable world model that can generate realistic ultrasound observations directly from clinical data remains largely unexplored.

We introduce SonoWorld, a causal, pose-controllable ultrasound world model that learns ultrasound video dynamics primarily from large-scale unlabeled clinical videos and requires only a small set of EM-tracked scans to acquire 6-DoF control. Inspired by previous works~\citep{gao2025adaworld,gao2026dreamdojo},  SonoWorld is trained in three stages: First, we pretrain a latent-action video diffusion model on around 6.9 million unlabeled multi-organ ultrasound frames, allowing it to capture both probe-induced appearance changes and physiological dynamics without pose annotations. Second, we ground the learned dynamics to physical probe motion by post-training on a small set of calibrated EM-tracked thyroid scans, where probe motion is represented as relative 6-DoF actions in the probe coordinate frame. Finally, we distill the bidirectional model into a causal autoregressive generator using self-forcing and selectively retain informative KV-cache tokens, enabling stable long-horizon interactive rollouts. Our main contributions are summarized as follows:
\begin{itemize}
    \item We introduce \textbf{SonoWorld}, to our knowledge the first causal, interactive, pose-controllable ultrasound world model that directly generates future ultrasound observations conditioned on 6-DoF probe actions, providing an interactive learned environment for robotic ultrasound data acquisition and policy evaluation.
    \item We develop a scalable training strategy that learns ultrasound dynamics from around 6.9 million unlabeled multi-organ clinical frames and grounds them to probe control using only a small set of EM-tracked scans. This enables large-scale dynamics learning without pose annotations, while requiring costly tracker-acquired data only for the subsequent physical action-grounding stage.
    \item We convert the bidirectional video model into a causal autoregressive generator through self-forcing and introduce selective KV-cache retention for streaming inference, enabling extended action-conditioned rollouts while preserving generation quality.
    \item We evaluate SonoWorld under measured, accelerated, round-trip, and axis-specific synthetic probe motions, and further conduct a user study with experienced ultrasound raters assessing image quality, video consistency, and probe--anatomy alignment.
\end{itemize}

\section{Related Work}
\paragraph{Generative World Models}
\label{sec:rw_world_models}
Generative world models learn action-conditioned dynamics of visual environments and have emerged as a scalable paradigm for interactive simulation across embodied AI~\citep{nvidia2025cosmos,yang2024unisim,jang2025dreamgen,gao2026dreamdojo,huang2026vid2world}, games~\citep{bruce2024genie,alonso2024diamond,kanervisto2025wham}, autonomous driving~\citep{hu2023gaia,gao2024vista,yang2025resim}. A central challenge in scaling such models is that action-labeled data are far less abundant than passive video. Latent-action models address this imbalance by discovering control-relevant dynamics directly from visual observations: Genie~\citep{bruce2024genie} learns discrete latent actions, AdaWorld~\citep{gao2025adaworld} extends them to a continuous action space, and subsequent work like DreamDojo scales this paradigm to real-world robotic application~\citep{gao2026dreamdojo} and grounds learned latent actions into executable robot commands~\citep{ye2025lapa}. This separation between learning visual dynamics from abundant unlabeled video and grounding them to physical actions provides a natural foundation for domains where paired action--video data are costly to acquire.
\paragraph{Medical World Models}
In the medical field generative world models have recently emerged as learned simulators for medical video. Endora~\citep{li2024endora} models spatiotemporal dynamics to generate realistic endoscopic videos, while Surgical Vision World Model~\citep{koju2025surgicalvision} learns latent actions from unlabeled surgical videos for controllable generation, and SAW~\citep{rapuri2026saw} further conditions laparoscopic video generation on language, tissue affordances, and 2D tool trajectories. In ultrasound, Cardiac Copilot~\citep{jiang2024cardiaccopilot} learns a latent spatial model from tracked echocardiography and predicts structural features of neighboring cardiac planes for probe guidance; Structure-aware World Model~\citep{jiang2024structure} extends this paradigm with large-scale self-supervised learning of anatomical representations under 3D pose transformations; and EchoWorld~\citep{yue2025echoworld} incorporates motion-induced visual dynamics and visual--motion history for sequential probe guidance, but these methods use world modeling primarily for representation learning rather than as image-generating interactive environments. More recently, Fan et al.~\citep{fan2026action} predict future ultrasound observations conditioned on tracked probe motions and use the learned dynamics for goal-plane navigation. In contrast, SonoWorld learns ultrasound dynamics primarily from large-scale untracked clinical videos, grounds them to metric 6-DoF probe actions using a small tracked subset, and causally generates future observations under these actions, enabling interactive rollouts for policy evaluation and model-based action discrimination.
\paragraph{Interactive Ultrasound Environments}
\label{sec:rw_environments}
Several strategies exist for building an environment in which probe motion can be explored.
Physical phantoms provide genuine acoustic interaction but lack clinical appearance and encode
a single fixed anatomy. Physics-based simulation renders images from a CT-derived anatomical
model at arbitrary poses~\citep{shams2008realtime, burger2013realtime, salehi2015patient}, yet
ray-casting cannot reproduce speckle from first principles~\citep{gao2009fast} and
wave-accurate solvers are too slow to be interactive~\citep{jaros2016full}. General-purpose robotics platforms such as NVIDIA Isaac Sim support scalable robotic ultrasound simulation and policy training, but their simplified ultrasound image formation does not faithfully capture the acoustic properties of clinical ultrasound~\citep{gao2026nvidia}. Sim-to-real translation narrows the
appearance gap~\citep{vitale2020improving} at the cost of physical consistency.
Freehand reconstruction compounds a tracked sweep into a volume that can be
re-sliced~\citep{lasso2014plus}, but view-dependent effects are frozen at the acquisition angle
and every simulated subject needs its own tracked acquisition. Scanning real patients is the
fidelity upper bound~\citep{su2024fully, jiang2025towards}, yet a scan traverses a single
trajectory and cannot report what a different motion would have produced.
A learned world model addresses these limitations, inheriting real appearance, view-dependent
acoustics and physiological motion from clinical video, covering anatomical variation across
patients, and answering potential pose queries without per-subject tracking.
\section{Methods}

\subsection{Foundational Ultrasound Video Diffusion Model}
\label{sec:foundational}
\paragraph{Ultrasound video dynamics encoding.}

Ultrasound video dynamics arise from both probe motion, which changes the imaged anatomical cross-section, and physiological motion, such as cardiac contraction, respiration, and vascular pulsation. To capture these dynamics without action annotations, we learn a latent action model (LAM) inspired by AdaWorld~\citep{gao2025adaworld}. Given a clip $\mathbf{x}^{1:K}$, the encoder patchifies each adjacent frame pair $(\mathbf{x}^i,\mathbf{x}^{i+1})$ into $16\times16$ patches, appends a learnable action token to each frame, and processes them with alternating spatial and temporal attention. The action token of $\mathbf{x}^{i+1}$ is compressed into a $32$-dimensional latent action $\mathbf{z}^i$, while the patch tokens are discarded. A decoder reconstructs $\mathbf{x}^{i+1}$ from $\mathbf{x}^i$ and $\mathbf{z}^i$. This low-dimensional bottleneck encourages $\mathbf{z}^i$ to encode frame-to-frame dynamics rather than appearance shared across frames. Full architectural details are provided in Appendix Fig~\ref{fig:lam_architecture}.
\begin{figure}[t]
\centering
\includegraphics[width=\linewidth]{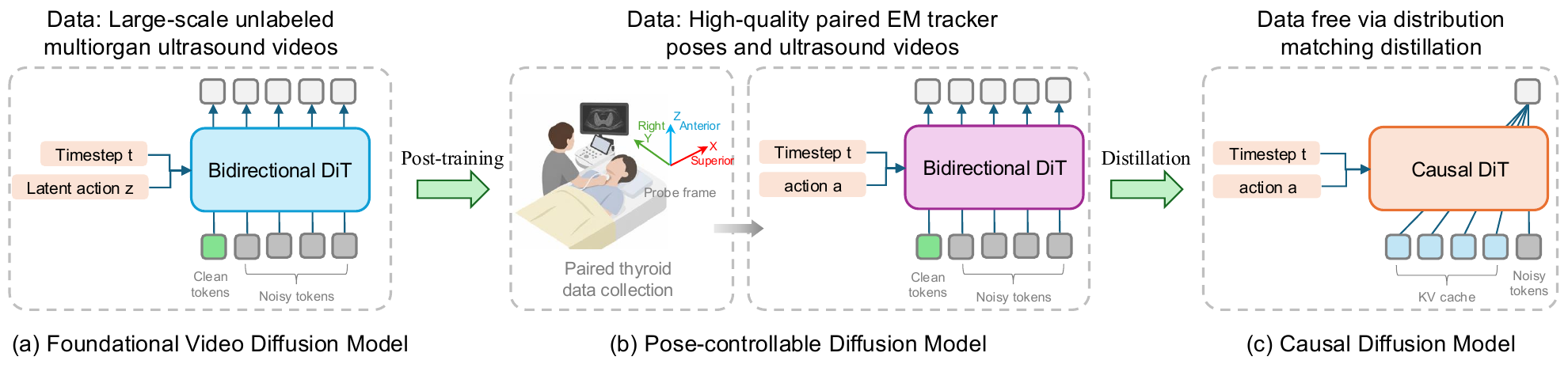}
\caption{Three stages of ultrasound world model development. (a) \textbf{Pretraining.} A bidirectional DiT is trained on large-scale unlabeled multi-organ ultrasound videos, conditioned on the diffusion timestep $t$ and latent actions $\mathbf{z}$ via adaptive layer normalization. (b) \textbf{Post-training.} The model is then post-trained on high-quality thyroid ultrasound videos paired with EM tracker poses, replacing the latent action with the physical action. (c) \textbf{Causal transform.} The bidirectional model is finally distilled into a causal DiT via distribution matching distillation (DMD)~\cite{yin2024one}.}
\label{fig:pipeline}
\end{figure}

\paragraph{Conditional video diffusion model training.}
We next train an action-conditioned image-to-world model that predicts future observations from an initial frame and a sequence of latent actions. We build on Wan2.1-T2V-1.3B~\citep{wan2025wan}, a latent video diffusion transformer, and encode each clip $\mathbf{x}^{1:K}$ into $\mathbf{y}=\mathcal{E}(\mathbf{x}^{1:K})$ using the pretrained Wan VAE $\mathcal{E}$. Generation is conditioned on the initial frame $\mathbf{x}^1$ and latent actions $\mathbf{z}^{1:K-1}$. We inject each $\mathbf{z}^i$ through adaptive layer normalization (AdaLN)~\citep{xu2019understanding} by projecting it to the model dimension, adding it to the diffusion timestep embedding, and mapping the result to per-block modulation parameters. The action projection is zero-initialized to preserve the pretrained video prior at the start of training. We optimize the model with flow matching~\citep{lipman2023flow,liu2023flow}. Given Gaussian noise $\boldsymbol{\epsilon}$ and $t\sim\mathcal{U}[0,1]$, we define $\mathbf{y}_t=(1-t)\mathbf{y}+t\boldsymbol{\epsilon}$ and train the velocity predictor $\mathbf{u}_\theta$ as
\begin{equation}
    \mathcal{L}_{\mathrm{flow}}(\theta)
    =
    \mathbb{E}_{\mathbf{y},\boldsymbol{\epsilon},\mathbf{c},t}
    \left[
        \left\|
        \mathbf{u}_\theta(\mathbf{y}_t,t,\mathbf{c})
        -
        (\boldsymbol{\epsilon}-\mathbf{y})
        \right\|_2^2
    \right],
\end{equation}
where $\mathbf{c}=(\mathbf{x}^1,\mathbf{z}^{1:K-1})$ denotes the initial-frame and latent-action conditions.
Video diffusion model training is detailed in Appendix~\ref{sec:implementation}.

\subsection{Pose-Controllable Video Diffusion Model}
\begin{figure}[t]
\centering
\includegraphics[width=\linewidth]{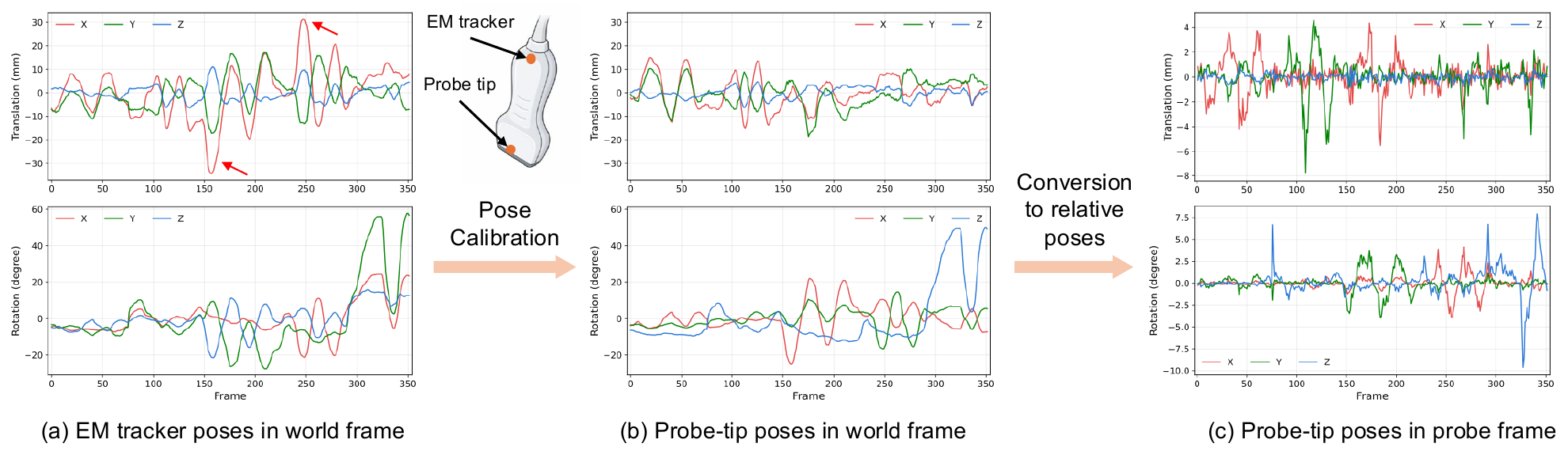}
\caption{
EM-tracked pose preprocessing pipeline (a) Raw EM sensor poses are expressed in world frame and are affected by the lever arm between the sensor and probe tip. (b) A fixed sensor-to-tip calibration transforms the measurements to probe-tip poses, such that the resulting motion corresponds to that of the imaging plane. (c) Consecutive probe-tip poses are converted to relative motions expressed in the preceding probe frame, yielding 6-DoF actions used to condition SonoWorld.}
\label{fig:pose_calibration}
\end{figure}

The latent actions learned during pretraining capture transition dynamics but lack physical units and therefore cannot directly control probe motion. We acquire freehand clinical scans with an EM tracker and calibrate the tracked sensor poses to the probe tip. We then express consecutive poses as relative 6-DoF motions in the probe frame (Fig.~\ref{fig:pose_calibration}) and replace the latent actions with these metric actions to post-train the pretrained model on the EM-tracked subset.
\label{sec:pose-controllable}

\paragraph{Pose calibration.}
The tracked sensor sits on the probe handle. The imaging plane is attached
to the tip, so a constant rigid offset, the lever arm, must be composed onto
every sample (Fig.~\ref{fig:pose_calibration}(a)(b)). Each scan records two synchronized streams: a sensor clamped
to the probe and a second sensor taped to the patient, which measures
patient motion. Each stream provides, at every sample $t$, a unit quaternion
$\mathbf{q}_t$, a position $\mathbf{p}_t \in \mathbb{R}^3$ in millimetres,
and a tracking quality $\kappa_t$, from which we form the homogeneous
transform $\mathbf{T}_{\text{world}\leftarrow\text{sensor}}(t) =
[\mathbf{R}(\mathbf{q}_t) \mid \mathbf{p}_t] \in SE(3)$, where
$\mathbf{T}_{B \leftarrow A}$ denotes the transform taking coordinates in
frame $A$ to frame $B$. The lever arm is obtained from the
manufacturer-supplied probe-to-sensor matrix by two signed axis
relabellings, yielding the constant transform
$\mathbf{T}_{\text{sensor}\leftarrow\text{tip}} = [\mathbf{R}_{\text{st}}
\mid \mathbf{r}]$, whose translation $\mathbf{r}$ is the tip position in
sensor coordinates ($\|\mathbf{r}\| \approx 119.2$\,mm for the thyroid
probe). Let $\mathbf{T}^{\text{probe}}_{\text{world}\leftarrow
\text{sensor}}(t)$ and $\mathbf{T}^{\text{ref}}_{\text{world}\leftarrow
\text{sensor}}(t)$ denote the probe- and patient-sensor streams. The probe-tip
pose in the patient frame factorizes as
\begin{equation}
  \mathbf{T}_{\text{patient}\leftarrow\text{tip}}(t) =
  \bigl(\mathbf{T}^{\text{ref}}_{\text{world}\leftarrow\text{sensor}}(t)\bigr)^{-1}\,
  \mathbf{T}^{\text{probe}}_{\text{world}\leftarrow\text{sensor}}(t)\,
  \mathbf{T}_{\text{sensor}\leftarrow\text{tip}} .
  \label{eq:tip_pose}
\end{equation}
Eq.~\eqref{eq:tip_pose} successively maps tip coordinates into the probe-sensor, world, and patient frames.
Derivation of pose calibration is
detailed in Appendix~\ref{sec:pose_calibration_apx}.
\paragraph{Conversion to relative poses.}
The action must describe the operator's motion rather than the orientation
and placement of the patient, so absolute world poses are unusable as
conditioning. The pose stream is therefore converted to relative poses
expressed in the probe's own frame. Abbreviating
$\mathbf{T}(t) \equiv \mathbf{T}_{\text{patient}\leftarrow\text{tip}}(t)$,
the relative pose between any ordered pair of samples $(s,t)$ is
\begin{equation}
  \mathbf{T}_{s\to t} = \mathbf{T}(s)^{-1}\,\mathbf{T}(t) =
  \begin{bmatrix}
    \mathbf{R}_s^{\top}\mathbf{R}_t & \mathbf{R}_s^{\top}(\mathbf{p}_t-\mathbf{p}_s)\\
    \mathbf{0}^{\top} & 1
  \end{bmatrix},
  \label{eq:rel_pose}
\end{equation}
where $\mathbf{R}_t$ and $\mathbf{p}_t$ are the rotation and translation
blocks of $\mathbf{T}(t)$, and $\mathbf{T}_{s\to t}$ denotes the motion of
the probe frame from sample $s$ to sample $t$ (Fig.~\ref{fig:pose_calibration}(c)). The patient frame
cancels between the two factors, so left-multiplying the whole stream by any
fixed transform leaves Equation.~\ref{eq:rel_pose} unchanged, and the
translation is resolved in the local axes of the start sample: advancing a
fixed distance along the beam is the same action wherever on the body it
occurs. Since $SE(3)$ is not a vector space, we map the relative pose to
$\mathbb{R}^6$ by concatenating its translation with the $\mathfrak{so}(3)$
logarithm of its rotation,
\begin{equation}
  \mathbf{a}_{s\to t} = \Bigl[\,\mathbf{R}_s^{\top}(\mathbf{p}_t-\mathbf{p}_s)
  \;\big\|\; \log\bigl(\mathbf{R}_s^{\top}\mathbf{R}_t\bigr)^{\vee}\Bigr]
  \in \mathbb{R}^6 ,
  \label{eq:action}
\end{equation}
with translation in millimetres and rotation in radians. 
\paragraph{Post-training with EM-tracked actions.}
We introduce a \emph{chunk action}, in which every frame carries its cumulative motion since the start of its own four-frame window, matching the temporal compression of the Wan VAE. Within each four-frame window, each frame is represented by its cumulative motion from the window start, and the four 6-DoF motions are concatenated into a $24$-dimensional action. Post-training is initialized from the pretrained model with the latent-action pathway replaced by a new metric-action pathway. Each chunk action is projected from $\mathbb{R}^{24}$ to the model dimension and added to the corresponding timestep embedding before adaptive layer normalization, with a zero action assigned to the initial latent frame. Generation is conditioned on the first clean latent frame, encoded as a prefix by concatenating the masked clean latents with a binary mask indicating observed frames and projecting them through a parallel convolution added to the backbone patch embedding. Both the action projection and prefix convolution are zero-initialized, preserving the pretrained function at initialization. The entire DiT generator is then trained end-to-end with all parameters unfrozen.

\subsection{Causal Video Diffusion Model}
\label{sec:causal}
The post-trained model uses bidirectional attention over all latent frames and therefore denoises an entire clip jointly, precluding interactive generation. An interactive probe simulator requires that the
next block of frames be emitted as soon as the operator supplies the next
action. We therefore convert the model into an efficient block-causal,
few-step generator that attends only backward in time and carries a KV cache
across blocks. Direct autoregressive training, however, introduces exposure bias~\citep{bengio2015scheduled,chen2024diffusion,liu2026rolling}: a model trained on ground-truth context must condition on its own predictions at inference, causing errors to accumulate over long rollouts. We therefore adopt self-forcing distillation~\citep{huang2026self}, in which each block is generated from a KV cache constructed from the student's own preceding outputs, reducing the mismatch between training and autoregressive inference.  Long-horizon generation additionally requires controlling the growth of the KV cache. Existing streaming video generators bound memory by retaining an attention sink and a recent temporal window while discarding intermediate history~\citep{liu2026rolling,lu2026reward}. We instead treat the choice of which frames to
retain as a selection problem. Inspired by heavy-hitter eviction in language
models~\citep{zhang2023h2o}, we score each cached frame by the attention
mass it has received and retain the highest-scoring frames alongside the
sink and the recent window, applying rotary embeddings when keys are written
into the cache so that retained frames keep their original positions. At inference, SonoWorld can generate ultrasound video at 40.6 fps at \(256\times256\) resolution using four denoising steps on a single NVIDIA H100, enabling interactive action-conditioned rollouts.

\section{Experiments}
\subsection{Datasets}
\paragraph{Unlabeled ultrasound videos.}
Under IRB approval, we collect a large-scale corpus of freehand ultrasound videos acquired
during routine clinical examinations, without any probe-pose annotation. The
raw corpus consists of 64{,}599 video files spanning three anatomical
regions: abdomen (31{,}209 files, 48.3\%), thyroid (21{,}399 files, 33.1\%),
and cardiac (11{,}991 files, 18.6\%), totaling 15{,}357{,}337 frames.
Because clinical recordings are noisy, we design an automated filtering
pipeline that removes colour Doppler acquisitions, discards static segments,
and excludes segments without probe--skin contact. Retained segments are
cropped to the imaging field of view, temporally resampled to 16\,fps,
resized to $256\times256$, and divided into non-overlapping 3-second chunks
of 45 frames each; segments shorter than 3 seconds are discarded. This
yields 153{,}892 training clips: abdomen (106{,}990), cardiac
(26{,}095), and thyroid (20{,}807), totaling 6,925,140 frames. These clips are used to train the latent-action video model and the foundational video diffusion
model. For validation, we hold out 500 clips: 64 thyroid clips together with 218 liver and 218 cardiac clips.
\paragraph{EM-tracked thyroid ultrasound videos.}
To condition the model on the probe motion, we collect thyroid ultrasound scans with synchronized 6-DoF pose measurements using a dual-sensor electromagnetic (EM) tracking system. One sensor is mounted on the probe handle and the second on the patient, allowing probe motion to be expressed relative to the subject rather than the field generator. Each scan follows a fixed acquisition protocol covering three translational and three rotational degrees of freedom. While actions are represented in full 6-DoF, surface-normal translation is physically constrained by probe--skin contact and tissue compression, so freehand scanning effectively explores a contact-constrained subset of the 6-DoF pose space. The dataset contains EM-tracked scans from 42 distinct subjects, split at the subject level into 33 for training and 9 for validation, with no subject overlap. Videos are cropped to the ultrasound field of view, resampled to 16\,fps, and resized to $256\times256$.  Each scan is cut into non-overlapping 45-frame windows, and every rollout is conditioned on the first latent frame of its window plus a 6-DoF action sequence. 
\subsection{Results on Ultrasound World Model Controllability}
To assess whether SonoWorld responds to commanded probe motion rather than merely extrapolating plausible video dynamics, we evaluate it under four complementary action settings. \textbf{Original pose} replays the measured probe trajectory and measures in-distribution action-following fidelity. \textbf{Zero motion} feeds an all-zero action sequence, so the rollout should stay on the initial frame with only physiological motion.  \textbf{Speed $\times2$} subsamples the trajectory by a factor of two, approximately doubling the displacement per step and testing whether the model responds to the commanded motion magnitude rather than a characteristic training-time velocity. \textbf{Round trip} follows the measured forward trajectory and then appends the exact inverse motion in reverse temporal order. This setting tests whether pose-conditioned generation remains stable over an extended trajectory that returns to the initial probe pose. For each setting, we quantify generation fidelity using PSNR, SSIM, and LPIPS against original sequence, reported as mean $\pm$ standard deviation over evaluation clips.

Table~\ref{tab:tasks} evaluates whether SonoWorld follows the commanded probe trajectory rather than merely producing plausible ultrasound dynamics. Under \textbf{Original pose}, the generated sequence is compared with the observations acquired along the measured trajectory; thus, PSNR, SSIM, and LPIPS quantify how accurately the model reproduces the image evolution induced by real probe motion. The causal model remains close to the bidirectional teacher, with only a 0.75 dB decrease in PSNR, a 0.016 decrease in SSIM, and a 0.026 increase in LPIPS, indicating that causal generation largely preserves pose-conditioned fidelity while enabling streaming inference with four denoising steps. \textbf{Zero motion} directly probes null-action stability: setting all 6-DoF actions to zero suppresses probe-induced evolution and all generated observations are substantially near the initial imaging plane, indicating that the generated dynamics respond to the supplied probe actions rather than following an autonomous motion prior. \textbf{Speed$\times2$} increases the probe displacement per generation step while preserving the underlying trajectory. Both models maintain nearly unchanged reconstruction quality, showing that the predicted image evolution scales with the commanded motion magnitude rather than a fixed temporal or motion prior. \textbf{Round trip} further tests whether the learned action representation is approximately reversible: after following the forward trajectory, the model is driven by the reversed actions in negated order. This is the most challenging setting, with PSNR decreasing by 0.43 dB for the bidirectional model and 0.92 dB for the causal model. The larger degradation of the causal model is consistent with error accumulation over the longer autoregressive rollout (23 versus 12 latent frames), while its retained fidelity indicates that learned control remains effective over extended bidirectional motion. We further evaluate axis-specific controllability using synthetic actions that vary one degree of freedom while holding the others fixed in Appendix Fig.~\ref{fig:gen_cust_pose}.

\begin{table}[t]
\centering
\caption{Quantitative controllability evaluation on EM-tracked scans. Ground-truth sequence is generated from original sequence according to the specific task. For zero motion it is the first image.}
\label{tab:tasks}
\small
\begin{tabular}{llccc}
\toprule
Models & Tasks & PSNR $\uparrow$ & SSIM $\uparrow$ & LPIPS $\downarrow$ \\
\midrule
\multirow{4}{*}{Bidirectional} & Original pose & 16.84$\pm$1.17 & 0.402$\pm$0.060 & 0.413$\pm$0.056 \\
                        & Zero motion   & 16.87$\pm$1.12 & 0.385$\pm$0.066 & 0.392$\pm$0.076 \\
                        & Speed$\times$2 & 16.89$\pm$0.98 & 0.414$\pm$0.053 & 0.411$\pm$0.051 \\
                        & Round trip    & 16.41$\pm$1.09 & 0.376$\pm$0.059 & 0.431$\pm$0.054 \\
\midrule
\multirow{4}{*}{Causal} & Original pose & 16.09$\pm$0.95 & 0.386$\pm$0.054 & 0.439$\pm$0.044 \\
                        & Zero motion   & 17.76$\pm$1.13 & 0.409$\pm$0.062 & 0.390$\pm$0.065 \\
                        & Speed$\times$2 & 16.10$\pm$1.05 & 0.388$\pm$0.051 & 0.445$\pm$0.048 \\
                        & Round trip    & 15.17$\pm$0.83 & 0.349$\pm$0.044 & 0.465$\pm$0.041 \\
\midrule
\end{tabular}
\end{table}

\begin{figure}[t]
\centering
\includegraphics[width=0.96\linewidth]{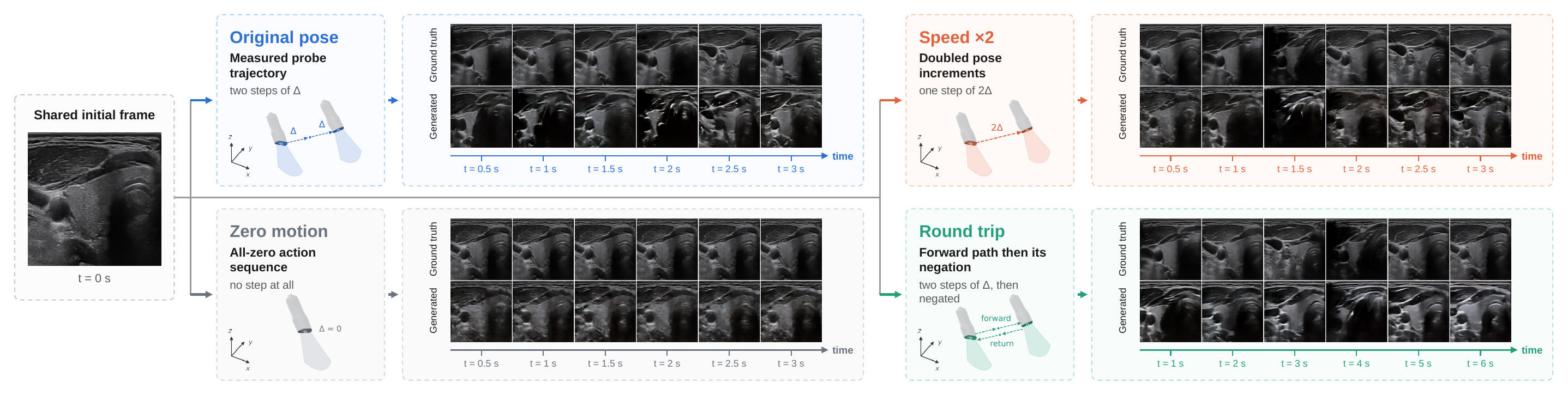}
\caption{Visualizations on video generation of our ultrasound world model. We utilize the causal model to autoregressively generate future frames conditioned on the same initial ultrasound frame (t=0s) and different probe trajectories. }
\label{fig:gen_org_pose}
\end{figure}

\subsection{Effect of Pretraining with Latent Actions}
We evaluate whether latent actions provide a effective conditioning signal for learning ultrasound dynamics from the large multi-organ corpus, where probe poses are unavailable. We compare two models with the same Wan2.1-T2V-1.3B backbone and training recipe, differing only in whether the learned 32-dimensional latent action is injected through adaptive layer normalization during training and inference. Evaluation is performed on an organ-balanced held-out set of 500 clips, where each rollout is conditioned on the first three latent frames and generates the remaining nine. As shown in Table~\ref{tab:latent_action}, latent-action conditioning improves PSNR by $3.60$ dB and SSIM by $0.083$, while reducing LPIPS by $0.067$. The substantial and consistent improvements across all three metrics demonstrate the effectiveness of latent-action conditioning in learning ultrasound video dynamics, showing that the learned latent actions capture transition-relevant information that substantially reduces the ambiguity of future observations from visual context alone.

\begin{table}[h]
\centering
\caption{Effect of latent actions on ultrasound video diffusion model pretraining.}
\label{tab:latent_action}
\small
\begin{tabular}{lccc}
\toprule
Methods & PSNR $\uparrow$ & SSIM $\uparrow$ & LPIPS $\downarrow$ \\
\midrule
w/o latent actions & 17.78$\pm$2.59          & 0.613$\pm$0.110          & 0.266$\pm$0.087 \\
w/ latent actions  & \textbf{21.38$\pm$2.80} & \textbf{0.696$\pm$0.108} & \textbf{0.199$\pm$0.072} \\
\bottomrule
\end{tabular}
\end{table}

\subsection{Ablation Studies on Causal Model Training Methods}
 
A practical interactive world model must generate observations causally as probe actions arrive, while remaining stable over autoregressive rollouts. Converting the bidirectional model to causal generation therefore introduces a central training challenge: errors in generated context can accumulate over time, and the choice of causal training strategy directly determines long-horizon rollout quality. We compare several causal training strategies under a controlled setting. All causal variants are initialized from the same post-trained bidirectional model, trained on the same 33 EM-tracked scans, and differ only in how the causal context is constructed during training. We evaluate 8-second rollouts (129 image frames; 33 latent frames) conditioned on an initial frame and the measured 6-DoF probe trajectory, with the bidirectional model included as a non-causal reference. As shown in Table~\ref{tab:ar_methods}, self-forcing~\citep{huang2026self} achieves the best perceptual quality among causal variants, with SSIM $0.394$ and LPIPS $0.436$, while remaining within $0.15$ dB PSNR of the bidirectional reference. Teacher forcing yields slightly higher PSNR ($16.44$ dB) but substantially worse LPIPS ($0.515$), while diffusion, rolling, and reward forcing degrade more consistently across metrics. These results indicate that exposing the model to its own generated context during training better matches the autoregressive inference distribution and preserves visual fidelity over long causal rollouts.

\begin{table}[h]
\centering
\caption{Comparison of causal model training methods on 8 second rollouts. For each metric, the
best result among causal methods is shown in boldface and the second best is
underlined.}
\label{tab:ar_methods}
\small
\begin{tabular}{lccc}
\toprule
Methods & PSNR $\uparrow$ & SSIM $\uparrow$ & LPIPS $\downarrow$ \\
\midrule
Bidirectional Model & 16.52$\pm$0.93 & 0.389$\pm$0.061 & 0.422$\pm$0.051 \\ \midrule
Teacher Forcing \citep{gao2024ca2}      & \textbf{16.44$\pm$0.76} & \underline{0.387$\pm$0.044} & 0.515$\pm$0.055 \\
Diffusion Forcing \citep{chen2024diffusion}    & 16.22$\pm$0.81 & 0.373$\pm$0.039 & 0.525$\pm$0.046 \\
Rolling Forcing \citep{liu2026rolling} & 15.77$\pm$0.90 & 0.372$\pm$0.055 & \underline{0.456$\pm$0.047} \\
Reward Forcing \citep{lu2026reward} & 15.60$\pm$0.81 & 0.371$\pm$0.051 & 0.469$\pm$0.041 \\
\midrule
\textbf{SonoWorld (Ours)} & \underline{16.37$\pm$0.98} & \textbf{0.394$\pm$0.052} & \textbf{0.436$\pm$0.047} \\
\bottomrule
\end{tabular}
\end{table}
\subsection{Ablation Studies on Causal Inference Techniques}
 
Long-horizon causal generation requires efficient management of temporal context. Retaining the full KV cache preserves all history but causes memory and attention cost to grow with rollout length, limiting scalable interactive inference. We therefore study whether a bounded cache can retain the information necessary for generation while reducing this cost. We apply different latent-frame-level KV-cache policies at inference time to the same self-forcing model, with no retraining, and evaluate 8-second rollouts containing 33 latent frames. \textbf{Full KV} retains all 33 frames (self-forcing default setting); \textbf{Local} keeps only the 12 most recent; \textbf{Sink+Local} retains 3 initial frames together with the 12 most recent~\citep{xiao2024efficient}; and \textbf{Eviction+Local} keeps 8 intermediate frames and the 4 most recent, where intermediate frames are selected either uniformly or by accumulated attention mass~\citep{zhang2023h2o}. As shown in Table~\ref{tab:kv_cache}, all bounded-cache variants remain within $0.14$ dB PSNR, $0.007$ SSIM, and $0.004$ LPIPS of full attention, despite retaining as few as 12 of 33 frames. SonoWorld with selective eviction achieves the best PSNR ($16.37$) and SSIM ($0.394$), while matching the best LPIPS ($0.436$). These results indicate that long-range context is useful but need not be retained densely: selecting informative historical frames can preserve generation quality while substantially reducing cache size and per-step attention cost.
\begin{table}[h]
\centering
\caption{Results on KV-cache configurations at inference time. We conducted causal rollouts of 8 second ultrasound videos conditioned on an initial frame and 6 DoF poses. 
}
\label{tab:kv_cache}
\small
\begin{tabular}{lccc}
\toprule
Cache Config. & PSNR $\uparrow$ & SSIM $\uparrow$ & LPIPS $\downarrow$ \\
\midrule
Full KV                 & 16.23$\pm$0.93 & 0.390$\pm$0.049 & 0.437$\pm$0.046 \\
Local                   & 16.26$\pm$0.86 & 0.387$\pm$0.049 & 0.440$\pm$0.045 \\
Sink+Local            & 16.31$\pm$0.96 & 0.391$\pm$0.051 & \textbf{0.436$\pm$0.048} \\
Eviction (strided)+Local & 16.27$\pm$0.94 & 0.391$\pm$0.051 & 0.437$\pm$0.047 \\
\midrule
\textbf{SonoWorld (Ours)}
& \textbf{16.37$\pm$0.98} & \textbf{0.394$\pm$0.052} & \textbf{0.436$\pm$0.047} \\
\bottomrule
\end{tabular}
\end{table}
\section{Applications of Ultrasound World Model}

\subsection{Interactive Application}
A practical ultrasound world model should provide plausible and anatomically consistent feedback under interactive probe control, enabling applications such as robotic policy evaluation, operator training, and exploration of probe--anatomy relationships. Because counterfactual probe trajectories often lack paired ground-truth videos, pixel-level reconstruction metrics alone are insufficient to assess such rollouts. We therefore conduct a user study with five experienced ultrasound raters on 55 generated clips, evaluating three complementary criteria: \textbf{image quality}, \textbf{video consistency}, and \textbf{probe--anatomy alignment}. We evaluate two settings. First, the initial frame is fixed while the probe follows three distinct trajectories (Path A--C), testing whether different actions induce coherent, path-dependent observations. Second, the probe trajectory is fixed while the conditioning frame is shifted laterally from the thyroid midline, with negative $\delta x$ corresponding to the left lobe and positive $\delta x$ to the right lobe. As shown in Table~\ref{tab:userstudy}, probe--anatomy alignment consistently receives the highest scores, followed by video consistency, while image quality receives relatively lower scores. These results indicate that SonoWorld maintains anatomically aligned and temporally consistent ultrasound observations across different probe trajectories and initial views, even when image fidelity is imperfect, which further confirms SonoWorld can closely follow the probe action.

\begin{table}[h]
\centering
\caption{\textbf{Human evaluation of pose-conditioned rollouts.}
Five experienced ultrasound raters scored 55 generated clips on three 5-point scales (1 = worst, 3=acceptable, 5 = perfect): image quality, video consistency, and probe--anatomy alignment. 
Values are mean $\pm$ 95\% CI over conditioning frames, after averaging raters within each clip.}
\label{tab:userstudy}
\small
\begin{tabular}{@{}l@{\hspace{0.9em}}lccc@{}}
\toprule
 & & Image quality & Video consistency & Probe--anatomy alignment \\
\midrule
\multicolumn{5}{l}{\emph{(a) Probe scan path varies; conditioning frame held fixed}} \\
& Path A & $3.15$\,$\pm$\,0.19 & $3.83$\,$\pm$\,0.26 & $4.08$\,$\pm$\,0.35 \\
& Path B & $3.35$\,$\pm$\,0.23 & $3.70$\,$\pm$\,0.36 & $3.73$\,$\pm$\,0.40 \\
& Path C & $3.60$\,$\pm$\,0.25 & $3.80$\,$\pm$\,0.13 & $3.92$\,$\pm$\,0.18 \\
\midrule
\multicolumn{5}{l}{\emph{(b) Conditioning frame shifted laterally; path held fixed}} \\
& $\delta x\!=\!-2$ (left lobe) & $3.58$\,$\pm$\,0.23 & $4.12$\,$\pm$\,0.12 & $4.35$\,$\pm$\,0.21 \\
& $\delta x\!=\!-1$ (left lobe) & $3.40$\,$\pm$\,0.33 & $3.85$\,$\pm$\,0.32 & $4.25$\,$\pm$\,0.21 \\
& $\delta x\!=\!0$ (middle) & $3.15$\,$\pm$\,0.19 & $3.83$\,$\pm$\,0.26 & $4.08$\,$\pm$\,0.35 \\
& $\delta x\!=\!+1$ (right lobe) & $3.12$\,$\pm$\,0.20 & $3.65$\,$\pm$\,0.08 & $3.48$\,$\pm$\,0.37 \\
& $\delta x\!=\!+2$ (right lobe) & $3.11$\,$\pm$\,0.28 & $3.60$\,$\pm$\,0.26 & $3.31$\,$\pm$\,0.43 \\
\bottomrule
\end{tabular}
\end{table}

\subsection{Model-Based Action Selection}
SonoWorld can support robotic navigation only if its predicted futures are sufficiently action-sensitive. We therefore evaluate the world model as an action-scoring mechanism for goal-directed navigation. Given an initial ultrasound frame and a target frame from a held-out tracked scan, we construct two equal-magnitude candidate trajectories along a single degree of freedom: one toward the target pose and one in the opposite direction. SonoWorld generates the terminal observation for each candidate, and PSNR to the target frame serves as the reward. A trial is successful when the action toward the true target receives the higher reward. Across held-out scans, SonoWorld achieves success rates of 74.8\% and 66.7\% at displacement thresholds of 15 and 20, respectively, indicating that its predicted observations contain action-dependent information that can be used to discriminate between candidate probe motions relative to a target view.
\begin{table}[h]
\centering
\caption{Results on model-based action discrimination. Entries are success rates (\%),
the fraction of rollouts in which the action reconstructs
the goal frame more accurately than its negation;}
\label{tab:full_negation}
\small
\setlength{\tabcolsep}{4pt}
\begin{tabular}{ccccccc}
\toprule
Threshold & $x$ & $y$ & $r_x$ & $r_y$ & $r_z$ & Avg. \\
\midrule
15 & 73.91 & 78.26 & 61.54 & 86.96 & 66.67 & 74.76 \\
20 & 88.89 & 77.78 & 66.67 & 77.78 & 44.44 & 66.67 \\
\bottomrule
\end{tabular}
\end{table}

\section{Conclusion}
We present SonoWorld, a causal, pose-controllable ultrasound world model that generates future observations conditioned on metric 6-DoF probe motion.  SonoWorld learns ultrasound dynamics from around 6.9 million unlabeled clinical ultrasound video frames, grounds them to physical probe actions using a small set of EM-tracked freehand scans, and is distilled into an autoregressive generator with bounded KV-cache inference. 
Although action grounding is evaluated on thyroid ultrasound, the SonoWorld framework is not organ-specific and can, in principle, be extended to other anatomical regions, such as abdominal organs, given appropriate tracked data for action grounding. Together, these results demonstrate the potential of SonoWorld as a data-driven interactive environment for robotic ultrasound acquisition and policy evaluation. It is worth noticing probe pose also does not fully determine ultrasound appearance: tissue deformation, physiological motion remain unobserved factors, indicating the limitation of this work. The absence of a shared public benchmark also prevents a direct quantitative comparison with the closest action-conditioned ultrasound world model, motivating standardized tracked-ultrasound benchmarks in future work.

\newpage

\bibliography{iclr2027_conference}
\bibliographystyle{iclr2027_conference}

\subsection*{AI use statement}
Generative AI assisted manuscript drafting, editing, literature and code inspection.

\subsection*{Ethics statement}
This study was conducted under institutional review board (IRB) approval and in accordance with the applicable requirements for research involving human data. Clinical ultrasound data were handled under the approved protocol and de-identified prior to research use. The EM-tracked ultrasound acquisition and expert evaluation were conducted under the corresponding approved procedures. We do not release identifiable clinical data, and all reported results are presented in aggregate.

\subsection*{Reproducibility statement}
We document data collection, splits and sampling, training settings, and evaluation protocols.

\appendix
\section{Appendix}
\subsection{Limitation of SonoWorld}
SonoWorld currently grounds physical actions using a relatively small thyroid-only EM-tracked cohort, and its learned dynamics are therefore validated only within the action and anatomical distributions represented by these data. Probe pose also does not fully determine ultrasound appearance: contact force, tissue deformation, physiological motion, and scanner settings remain unobserved factors. Pixel-level metrics such as PSNR, SSIM, and LPIPS provide only indirect evidence of geometric control, and null-action rollouts exhibit. In addition, the lack of a shared tracked-ultrasound benchmark prevents a direct quantitative comparison with the closest action-conditioned ultrasound world models. SonoWorld should therefore be viewed as a learned, data-driven environment rather than a physically guaranteed ultrasound simulator.

\subsection{Detailed Derivation for Pose Calibration and Relative Pose Conversion}
\label{sec:pose_calibration_apx}

This appendix expands Eqs.~\eqref{eq:tip_pose}--\eqref{eq:action}. Throughout,
$t$ indexes tracker samples rather than diffusion time.

\paragraph{Notation and pose convention.}
Each scan records two synchronized EM streams: a sensor clamped to the probe
handle and a second sensor taped to the patient, whose only purpose is to
measure patient motion. Each stream provides, per sample $t$, a timestamp, a
unit quaternion $\mathbf{q}_t = (w,x,y,z)$, a position
$\mathbf{p}_t\in\mathbb{R}^3$ in millimetres, and a scalar tracking quality
$\kappa_t$. All geometry uses homogeneous $4\times4$ transforms in the
canonical column-vector convention,
\begin{equation}
  \mathbf{T} = \begin{bmatrix} \mathbf{R} & \mathbf{p}\\
  \mathbf{0}^{\top} & 1\end{bmatrix} \in SE(3),
  \qquad
  \tilde{\mathbf{x}}_{\text{target}} = \mathbf{T}\,\tilde{\mathbf{x}}_{\text{source}},
  \qquad
  \mathbf{T}^{-1} = \begin{bmatrix} \mathbf{R}^{\top} & -\mathbf{R}^{\top}\mathbf{p}\\
  \mathbf{0}^{\top} & 1\end{bmatrix},
  \label{eq:se3}
\end{equation}
with $\tilde{\mathbf{x}} = [\mathbf{x}^{\top},1]^{\top}$ and the product
$\mathbf{T}_A\mathbf{T}_B$ meaning $A$ applied after $B$. We write
$\mathbf{T}_{B\leftarrow A}$ for the transform taking coordinates in frame $A$
to frame $B$, so that transforms compose by matching adjacent subscripts,
$\mathbf{T}_{C\leftarrow A} = \mathbf{T}_{C\leftarrow B}\mathbf{T}_{B\leftarrow A}$.

The mapping from the stored quaternion to a rotation matrix carries two binary
ambiguities: whether $w$ is stored first or last, and whether $\mathbf{R}$ maps
sensor to world or world to sensor. We measure rather than assume them. All
four combinations are scored by how well the sensor origin traces a sphere
about a fixed pivot, and the winner is decided by majority vote over up to
eight scans, since the layout is a property of the recording software and
cannot vary per scan. On this corpus the detected convention is
$w$-first with a transpose, that is
\begin{equation}
  \mathbf{x}_{\text{world}} = \mathbf{R}(\mathbf{q}_t)\,\mathbf{x}_{\text{sensor}}
  + \mathbf{p}_t,
  \qquad
  \mathbf{T}_{\text{world}\leftarrow\text{sensor}}(t)
  = \bigl[\mathbf{R}(\mathbf{q}_t)\mid\mathbf{p}_t\bigr].
  \label{eq:convention}
\end{equation}

\paragraph{From sensor pose to probe-tip pose.}
The tracked sensor sits on the probe handle, roughly $119$\,mm from the
acoustic tip, whereas the imaging plane is attached to the tip. The two frames
are rigidly connected, so a constant offset, the lever arm, must be composed
onto every sample. The manufacturer supplies a probe-to-sensor matrix
$\mathbf{M}_{\text{P2S}}$; converting it into the frame our files use requires
two signed axis relabellings,
\begin{equation}
  \mathbf{T}_{\text{sensor}\leftarrow\text{tip}} =
  \mathbf{A}_{\text{file}\leftarrow\text{vendor}}\;
  \mathbf{M}_{\text{P2S}}\;
  \mathbf{A}_{\text{tip}\leftarrow\text{probe}}^{-1},
  \label{eq:calib}
\end{equation}
where each $\mathbf{A}$ is a signed permutation matrix with determinant $+1$,
so that the relabelling stays right-handed. For the thyroid probe,
$\mathbf{A}_{\text{file}\leftarrow\text{vendor}}:(x,y,z)\mapsto(-Y,-Z,+X)$,
selected by scoring all $48$ signed axis permutations against the tracking
data, and
$\mathbf{A}_{\text{tip}\leftarrow\text{probe}}:(x,y,z)\mapsto(-Y,+Z,-X)$, which
names the tip frame as (transverse, transverse, axial) with the axial axis
pointing from the sensor toward the tip. Writing
$\mathbf{T}_{\text{sensor}\leftarrow\text{tip}} = [\mathbf{R}_{\text{st}}\mid\mathbf{r}]$,
the translation $\mathbf{r}$ is the tip position in sensor coordinates, with
$\|\mathbf{r}\|\approx119.2$\,mm for the thyroid probe, and
$\mathbf{R}_{\text{st}}$ is the probe frame. The delivered thyroid matrix
carries a $1.7^{\circ}$ tilt away from an exact axis permutation, so it is a
measured calibration rather than a CAD ideal; the cardiac and abdomen probes
are exact $90^{\circ}$ turns about the sensor $z$ axis.

\paragraph{Two-sensor composition.}
Let $\mathbf{T}^{\text{probe}}_{\text{world}\leftarrow\text{sensor}}(t)$ and
$\mathbf{T}^{\text{ref}}_{\text{world}\leftarrow\text{sensor}}(t)$ be the probe-
and patient-sensor poses in the world frame of the field generator. The
quantity of interest is the tip pose expressed in the patient frame, which
factorizes uniquely as Eq.~\eqref{eq:tip_pose},
\begin{equation}
  \mathbf{T}_{\text{patient}\leftarrow\text{tip}}(t) =
  \bigl(\mathbf{T}^{\text{ref}}_{\text{world}\leftarrow\text{sensor}}(t)\bigr)^{-1}\,
  \mathbf{T}^{\text{probe}}_{\text{world}\leftarrow\text{sensor}}(t)\,
  \mathbf{T}_{\text{sensor}\leftarrow\text{tip}} .
\end{equation}
The order is forced by the frame in which each factor is written:
$\mathbf{T}_{\text{sensor}\leftarrow\text{tip}}$ multiplies on the right
because it is expressed in the probe-sensor frame, and
$(\mathbf{T}^{\text{ref}}_{\text{world}\leftarrow\text{sensor}})^{-1}$
multiplies on the left because it changes the world frame. Abbreviating the
blocks as $[\mathbf{R}_{\text{probe},t}\mid\mathbf{p}_{\text{probe},t}]$ and
$[\mathbf{R}_{\text{ref},t}\mid\mathbf{p}_{\text{ref},t}]$ and expanding with
Eq.~\eqref{eq:se3},
\begin{equation}
  \mathbf{T}^{\text{probe}}_{\text{world}\leftarrow\text{sensor}}(t)\,
  \mathbf{T}_{\text{sensor}\leftarrow\text{tip}}
  = \bigl[\,\mathbf{R}_{\text{probe},t}\mathbf{R}_{\text{st}}
    \;\big|\; \mathbf{R}_{\text{probe},t}\mathbf{r} + \mathbf{p}_{\text{probe},t}\,\bigr],
\end{equation}
and left-multiplying by
$[\mathbf{R}_{\text{ref},t}^{\top}\mid-\mathbf{R}_{\text{ref},t}^{\top}\mathbf{p}_{\text{ref},t}]$
gives
\begin{equation}
  \mathbf{R}^{\text{tip}}_t = \mathbf{R}_{\text{ref},t}^{\top}\,
  \mathbf{R}_{\text{probe},t}\,\mathbf{R}_{\text{st}},
  \qquad
  \mathbf{p}^{\text{tip}}_t = \mathbf{R}_{\text{ref},t}^{\top}
  \bigl(\mathbf{p}_{\text{probe},t} + \mathbf{R}_{\text{probe},t}\mathbf{r}
  - \mathbf{p}_{\text{ref},t}\bigr).
  \label{eq:tip_expanded}
\end{equation}
Both rotations participate: the reference rotation acts on the whole
probe-to-reference vector, not only on the tip's orientation. Omitting
$\mathbf{R}_{\text{ref},t}^{\top}$ from the translation, which is the usual
form of this error, incurs a discrepancy of the reference rotation times the
inter-sensor distance, approximately $130$\,mm on this data.

A distant patient sensor injects more error than the patient motion it
removes, because its orientation error is multiplied by the inter-sensor
baseline and lies almost entirely below $0.5$\,Hz, so no zero-phase filter
removes it. Reference compensation is therefore applied only when the mean
baseline satisfies $\bar{b}\le 300$\,mm, and otherwise the tip is left in the
world frame,
\begin{equation}
  \mathbf{T}^{\text{tip}}_t =
  \begin{cases}
    \bigl(\mathbf{T}^{\text{ref}}_{\text{world}\leftarrow\text{sensor}}(t)\bigr)^{-1}
    \mathbf{T}^{\text{probe}}_{\text{world}\leftarrow\text{sensor}}(t)\,
    \mathbf{T}_{\text{sensor}\leftarrow\text{tip}}, & \bar{b}\le 300\,\text{mm},\\[4pt]
    \mathbf{T}^{\text{probe}}_{\text{world}\leftarrow\text{sensor}}(t)\,
    \mathbf{T}_{\text{sensor}\leftarrow\text{tip}}, & \text{otherwise.}
  \end{cases}
  \label{eq:ref_gating}
\end{equation}
On the 42 retained EM thyroid segments this yields 35 patient-referenced and 7
world-frame segments.

\paragraph{World-frame poses to egocentric relative poses.}
The action must describe what the operator does rather than the orientation
and placement of the patient, so absolute world poses are unusable as
conditioning: they are not invariant to the transmitter frame. Abbreviating
$\mathbf{T}(t)\equiv\mathbf{T}_{\text{patient}\leftarrow\text{tip}}(t) =
[\mathbf{R}_t\mid\mathbf{p}_t]$, the pose stream is converted to relative poses
expressed in the probe's own frame. For any ordered pair of samples $(s,t)$,
Eq.~\eqref{eq:se3} gives
\begin{equation}
  \mathbf{T}_{s\to t} = \mathbf{T}(s)^{-1}\mathbf{T}(t)
  = \begin{bmatrix} \mathbf{R}_s^{\top} & -\mathbf{R}_s^{\top}\mathbf{p}_s\\
    \mathbf{0}^{\top} & 1\end{bmatrix}
    \begin{bmatrix} \mathbf{R}_t & \mathbf{p}_t\\
    \mathbf{0}^{\top} & 1\end{bmatrix}
  = \begin{bmatrix} \mathbf{R}_s^{\top}\mathbf{R}_t &
    \mathbf{R}_s^{\top}(\mathbf{p}_t-\mathbf{p}_s)\\
    \mathbf{0}^{\top} & 1\end{bmatrix},
  \label{eq:rel_pose_apx}
\end{equation}
which is Eq.~\eqref{eq:rel_pose}. Since $SE(3)$ is not a vector space, the
6-DoF action is the concatenation of the translation block with the rotation
vector, that is the axis-angle vector obtained from the $\mathfrak{so}(3)$
logarithm of the rotation block. For $\mathbf{R}\in SO(3)$ with angle
$\theta = \arccos\bigl((\mathrm{tr}\,\mathbf{R}-1)/2\bigr)$,
\begin{equation}
  \log(\mathbf{R}) = \frac{\theta}{2\sin\theta}\bigl(\mathbf{R}-\mathbf{R}^{\top}\bigr),
  \qquad
  \begin{bmatrix} 0 & -\omega_3 & \omega_2\\ \omega_3 & 0 & -\omega_1\\
  -\omega_2 & \omega_1 & 0\end{bmatrix}^{\vee} = \boldsymbol{\omega},
  \label{eq:logmap}
\end{equation}
so that $\log(\mathbf{R})^{\vee} = \theta\mathbf{u}$ with $\mathbf{u}$ the unit
rotation axis. Combining Eqs.~\eqref{eq:rel_pose_apx} and~\eqref{eq:logmap}
gives Eq.~\eqref{eq:action},
\begin{equation}
  \mathbf{a}_{s\to t} = \bigl[\,
  \underbrace{\mathbf{R}_s^{\top}(\mathbf{p}_t-\mathbf{p}_s)}_{\text{local translation, mm}}
  \;\big\|\;
  \underbrace{\log\bigl(\mathbf{R}_s^{\top}\mathbf{R}_t\bigr)^{\vee}}_{\text{relative rotation, rad}}
  \,\bigr] \in \mathbb{R}^6 .
\end{equation}
Two properties matter. First, left-multiplying the whole stream by any fixed
world transform, $\mathbf{T}(\cdot)\mapsto\mathbf{G}\mathbf{T}(\cdot)$, leaves
$(\mathbf{G}\mathbf{T}(s))^{-1}\mathbf{G}\mathbf{T}(t) =
\mathbf{T}(s)^{-1}\mathbf{T}(t)$ unchanged, so the action is invariant to the
choice of world frame. Second, the translation is resolved in the local axes of
the start anchor, which makes it egocentric: advancing $2$\,mm along the beam
is the same action wherever on the body it occurs. The rotation vector is used
rather than a quaternion or Euler angles because it is a continuous, minimal,
three-dimensional parameterization with no gimbal lock and no double cover,
which is what a regression target requires.

\paragraph{Chunk-anchored action.}
The Wan VAE compresses time by $4\times$ with an independent first frame, so
the $45$ pixel frames of a clip map to $12$ latent frames, with latent frame
$i$ anchored at pixel frame $\tau_i = 4i$ for $i = 0,\dots,11$. Taking the net
motion of each latent frame, $\mathbf{a}_{\tau_{i-1}\to\tau_i}$, would discard
the shape of the motion inside it: a smooth $2$\,mm sweep and a jerk followed
by a stop produce the same net displacement. We therefore anchor every pixel
frame to the start of its own window instead. For each pixel frame
$f = 1,\dots,44$, define its chunk anchor and the corresponding action
\begin{equation}
  \pi(f) = 4\Bigl\lfloor \frac{f-1}{4}\Bigr\rfloor \in \{0,4,8,\dots,40\},
  \qquad
  \mathbf{a}'_f = \mathbf{a}_{\pi(f)\to f},
  \label{eq:chunk_action}
\end{equation}
so that every frame carries the cumulative motion since the start of its own
four-frame window,
\begin{equation}
  \underbrace{0\!\to\!1,\;0\!\to\!2,\;0\!\to\!3,\;0\!\to\!4}_{\text{latent frame }1},\quad
  \underbrace{4\!\to\!5,\;4\!\to\!6,\;4\!\to\!7,\;4\!\to\!8}_{\text{latent frame }2},\quad
  \underbrace{8\!\to\!9,\;\dots}_{\text{latent frame }3},\;\dots
\end{equation}
yielding $\mathbf{A}'\in\mathbb{R}^{44\times6}$ with no null row. Grouping the
four entries belonging to each latent frame gives $\mathbb{R}^{11\times24}$,
and a zero row is prepended inside the generator for latent frame $0$, which
has no predecessor, restoring the $12$-step alignment. Each group traces the
within-chunk trajectory $0\to1\to2\to3\to4$ rather than only its endpoint,
while its final entry $\mathbf{a}'_{4i}$ is the net motion
$\mathbf{a}_{\tau_{i-1}\to\tau_i}$ of that latent frame, so no information is
lost relative to the one-vector-per-latent-frame encoding.

\paragraph{Normalization and quality weighting.}
Translation is in millimetres and rotation in radians, quantities that differ
by orders of magnitude, so each of the six dimensions is standardized with
statistics fitted on the training split only and applied unchanged to
evaluation,
\begin{equation}
  \tilde{\mathbf{a}}^{(j)} = \frac{\mathbf{a}^{(j)} - \mu_j}{\sigma_j},
  \qquad \sigma_j \leftarrow \max(\sigma_j, 10^{-6}).
  \label{eq:action_norm}
\end{equation}
The statistics are fitted over all $44$ rows, none of which is a structural
zero. Each action step additionally carries a pooled tracking quality
$\bar{\kappa}_i$, the mean of $\kappa$ over the frames spanned by that step,
and a binary weight $w_i$, set to $1$ when $\bar{\kappa}_i < 2000$ and to $0$
otherwise, with missing quality treated as usable. Training can therefore mask
the steps recorded under weak tracking.

\subsection{Latent Action Model Training}
\begin{figure}[h]
\centering
\includegraphics[width=0.7\linewidth]{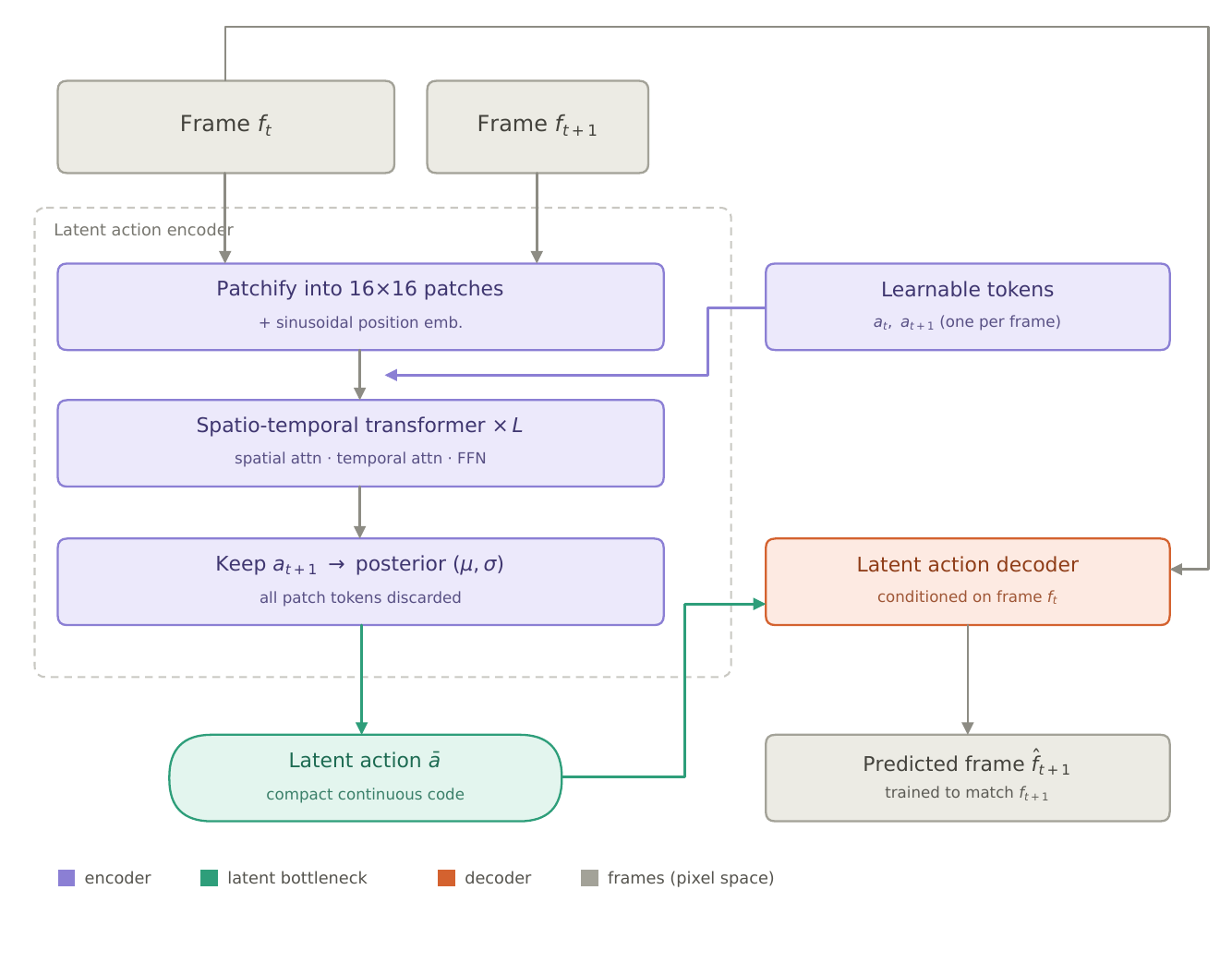}
\caption{Network architecture of the latent action model.}
\label{fig:lam_architecture}
\end{figure}

\paragraph{Architecture of Latent Action Model.}
The latent action model (LAM)~\citep{gao2025adaworld} is a conditional variational autoencoder that infers a compact code for the transition between two frames (Fig.~\ref{fig:lam_architecture}).
Given a source frame $x^{i}$ and a target frame $x^{i+1}$, each frame is split into non-overlapping patches, giving patch-token sequences $u^{i}$ and $u^{i+1}$. A learnable action token, $z^{i}$ or $z^{i+1}$, is prepended to each sequence.
The encoder $E_\phi$ is a spatio-temporal transformer whose blocks alternate spatial attention within each frame with temporal attention across the two frames.
Through temporal attention, the action token of the target frame gathers information from the source frame. We keep only this token and discard all patch tokens.
A linear head maps this token to the mean $\mu$ and variance $\sigma^{2}$ of a diagonal Gaussian posterior $q_\phi(z^{i+1}\mid x^{i},x^{i+1})$, which yields a low-dimensional latent action $z^{i+1}$.
The decoder $D_\psi$ is a spatial-only transformer. It predicts the target frame $\hat{x}^{i+1}$ from the source tokens $u^{i}$, with $z^{i+1}$ added to every token.
Because the decoder never sees $x^{i+1}$, the target frame can reach its own reconstruction only through $z^{i+1}$.
The source frame already provides the appearance, so this bottleneck forces $z^{i+1}$ to encode the change between the two frames (probe motion and the resulting anatomical displacement) rather than the anatomy itself.

\paragraph{Training details.}
Training pairs are sampled from clips resampled to $16$\,fps. The source frame $x^{i}$ is drawn uniformly from a clip, and the target frame $x^{i+1}$ is taken $\delta\sim\mathcal{U}\{1,\dots,4\}$ frames later, so that $z^{i+1}$ captures probe motions of variable duration. Both frames of a pair receive the same random brightness shift.
During training, $z^{i+1}$ is sampled with the reparameterization trick, $z^{i+1}=\mu+\sigma\odot\epsilon$ with $\epsilon\sim\mathcal{N}(0,I)$. At inference, we use the posterior mean $z^{i+1}=\mu$. The training objective combines a per-pixel reconstruction loss with a KL regularizer toward a standard normal prior:
\begin{equation}
  \mathcal{L}(\phi,\psi)=
  \mathbb{E}\bigl[\|x^{i+1}-\hat{x}^{i+1}\|_2^2\bigr]
  +\beta\,D_{\mathrm{KL}}\bigl(q_\phi(z^{i+1}\mid x^{i},x^{i+1})\,\|\,\mathcal{N}(0,I)\bigr).
\end{equation}
We set $\beta$ small, so the KL term only keeps the posterior scales well-conditioned. The bottleneck therefore comes mainly from the low dimensionality of $z^{i+1}$ and from the decoder's lack of access to $x^{i+1}$. We monitor PSNR and SSIM of $\hat{x}^{i+1}$ as reconstruction diagnostics. Table~\ref{tab:lam_hparams} lists all hyperparameters.

\begin{table}[t]
  \centering
  \caption{Hyperparameters of the latent action model.}
  \label{tab:lam_hparams}
  \small
  \begin{tabular}{ll}
    \toprule
    Hyperparameter & Value \\
    \midrule
    Frame resolution / patch size $p$ & $256\times256$ / $16$ \\
    Patch tokens per frame / token dim. & $256$ / $768$ \\
    Width / attention heads & $1024$ / $16$ (head dim.\ $64$) \\
    Encoder / decoder blocks & $24$ / $24$ \\
    Positional encoding & sinusoidal (spatial), RoPE (temporal) \\
    Latent action dim.\ $d_z$ & $32$ \\
    KL weight $\beta$ & $10^{-6}$ \\
    Frame gap $\delta$ & $\mathcal{U}\{1,\dots,4\}$ at $16$\,fps \\
    Brightness jitter & $\mathcal{U}(-0.1,\,0.1)$, shared within a pair \\
    Optimizer & AdamW \\
    Learning rate / weight decay & $2.5\times10^{-5}$ / $10^{-2}$ \\
    Gradient clipping & $0.3$ \\
    Precision & bf16 \\
    Batch size & $128$/GPU $\times$ $4$ GPUs $=512$ pairs \\
    Training corpus & $153{,}892$ ultrasound clips \\
                    & (abdomen $69.5\%$ / cardiac $17\%$ / thyroid $13.5\%$) \\
    \bottomrule
  \end{tabular}
\end{table}

\subsection{Implementation details}
\label{sec:implementation}

\paragraph{Pretraining data.}
The foundational model is trained on the full 153{,}892-clip multi-organ
corpus, composed of $69.5\%$ abdomen, $17.0\%$ cardiac and $13.5\%$ thyroid
clips. Training reads cached latents and cached actions rather than pixels:
each clip is stored as a $12\times16\times32\times32$ latent, the Wan VAE
encoding of 45 frames at $256\times256$, together with its $12\times32$
sequence of latent actions, and a per-organ T5 text embedding cached once and
looked up by organ label. The VAE, the text encoder and the latent action
model are therefore absent from the training loop, which is what makes a
768-clip effective batch affordable. No data augmentation is applied at this
stage: at this corpus size the data supplies its own diversity, and caching is
the more valuable trade.

\paragraph{Post-training data and temporal augmentation.}
The EM-tracked corpus is three orders of magnitude smaller, 42 tracked thyroid
scans yielding 262 clips, which is too few to learn a controllable mapping
from a 24-dimensional action to nine generated latent frames. We therefore
invert the trade made above and do not train on precomputed clips at all. The
dataloader reads whole segments, samples a 45-frame window inside the
probe-contact interval, applies one temporal augmentation, and VAE-encodes the
result on the fly. Every augmentation is expressed as a length-45 integer
index array into the source frames; the same index is applied to the pixel
frames and to the per-frame tip pose, and the action is recomputed from the
indexed pose by the procedure of Section~\ref{sec:pose_calibration_apx}. No
per-augmentation action algebra is needed, and the action is correct by
construction for any index map. Table~\ref{tab:augmentation} lists the four
cases, sampled with equal probability.

\begin{table}[h]
\centering
\caption{Online temporal augmentation used in post-training. Each case is an
index map into the source segment; the action is recomputed from the indexed
pose rather than edited.}
\label{tab:augmentation}
\small
\setlength{\tabcolsep}{6pt}
\begin{tabular}{llll}
\toprule
Augmentation & Prob. & Index map & Effect on the action \\
\midrule
Identity    & 0.25 & $[s,\,s+45)$ & unchanged \\
Reverse     & 0.25 & $[s+44,\,s]$ descending & motion played backward \\
Stride      & 0.25 & $s + \sigma[0,\dots,44]$, $\sigma\in\{2,3\}$ & composed, faster motion \\
Stop-motion & 0.25 & monotonic non-decreasing with repeats & exact zero-motion steps \\
\bottomrule
\end{tabular}
\end{table}

Stop-motion freezes a fraction $\rho\sim\mathcal{U}(0.1,0.35)$ of the 45 output
steps by repeating the current frame, which yields action entries that are
exactly zero. This is the only mechanism that places a null action inside the
training distribution, and it is what teaches the model the correspondence
between a zero action and a static image. Stride augmentation requires a
longer source window, 133 frames at $\sigma = 3$, which is why segments rather
than pre-cut clips are the unit of storage. Because windows and augmentations
are resampled online, a pass over the 262 slots does not repeat data: the
effective sample space is the set of draws of window start, augmentation and
noise.

\paragraph{Action control via adaptive layer normalization.}
The action enters the network only through the linear embedding
$\mathbf{a}_i = W_a\mathbf{u}_i + \mathbf{b}_a$ with $W_a \in \mathbb{R}^{d \times d_a}$.
Everything downstream of it (the shared projection, the per-block offsets, all transformer
blocks and the head) operates at width $d$ and does not depend on the action dimension $d_a$.
Switching action spaces therefore requires replacing only this one matrix.
During pretraining, $\mathbf{u}_i$ is the 32-dimensional LAM latent action of latent frame $i$,
so $d_a = 32$.
During post-training, each clip of 45 pixel frames provides 44 per-frame 6-DoF motions, 
one for every pixel frame after the first, each measured from the start
of its four-frame window. The four motions that fall within the same latent frame are
concatenated into a single 24-dimensional chunk action, which yields 11 chunk actions,
and latent frame~0, which has no predecessor, is assigned a zero vector.
This gives one action per latent frame for all 12 latent frames, with $d_a = 24$.
When initializing post-training from the pretrained checkpoint, we discard the pretrained
$1536 \times 32$ action embedding and allocate a new $1536 \times 24$ embedding with zero
weights and bias, so that the new action pathway is learned from scratch while all other
parameters are inherited.
With all parameters unfrozen, the 24-dimensional embedding is learned quickly during
end-to-end post-training.

\paragraph{Training objective.}
Both stages use the same objective. The primary term is the rectified-flow
loss of the backbone, with 1000 training timesteps and a timestep shift of
$5.0$, weighted by the scheduler timestep weighting $w(t)$ and masked to the
generated frames so that the given prefix is never supervised,
\begin{equation}
  \mathcal{L}_{\text{flow}} = \frac{\sum_{b,i} m_{b,i}\,w(t_{b,i})\,
  \bigl\|\mathbf{v}_\theta - \mathbf{v}^{\star}\bigr\|^2_{b,i}}
  {\sum_{b,i} m_{b,i}},
  \qquad m_{b,i} = 1 \text{ for } i \ge k,\; 0 \text{ otherwise},
\end{equation}
where $\mathbf{v}_\theta$ is the predicted velocity, $\mathbf{v}^{\star}$ its
target, and $k = 3$ the number of given prefix frames. A per-frame loss of
this form is largely satisfied by a model that renders each frame plausibly
while ignoring the action, since most of the latent energy sits in static
anatomy. We therefore add a motion-consistency term that matches the
frame-to-frame change of the prediction to that of the target,
\begin{equation}
  \mathcal{L}_{\text{motion}} = \frac{\sum_{b,i} m_{b,i}\,w(t_{b,i})\,
  \bigl\|\Delta_i\mathbf{v}_\theta - \Delta_i\mathbf{v}^{\star}\bigr\|^2}
  {\sum_{b,i} m_{b,i}},
  \qquad \Delta_i\mathbf{v} = \mathbf{v}_i - \mathbf{v}_{i-1},
\end{equation}
masked to transitions ending in a generated frame, so that the handoff from
the prefix to the first generated frame is included. This term measures only
what changes between frames, which is precisely what the action controls. The
total objective is
\begin{equation}
  \mathcal{L} = \mathcal{L}_{\text{flow}} + \lambda\,\mathcal{L}_{\text{motion}},
  \qquad \lambda = 0.1 .
\end{equation}
$\mathcal{L}_{\text{motion}}$ is computed and logged whether or not it is
weighted, so the setting $\lambda = 0$ is a logging-identical control.

\paragraph{Optimization.}
Pretraining follows a four-stage curriculum in which each stage warm-starts
the next and every stage trains the full network. A per-frame stage on single
latent frames teaches speckle statistics and anatomy without any temporal
burden, at an effective batch of 1536 frames; a video stage extends to whole
12-frame clips; a prefix stage adds the three-frame conditioning branch; and
the final stage adds the latent-action branch. The last three stages all use
an effective batch of 768 clips, so one epoch over the corpus is 201 optimizer
steps. Post-training warm-starts from the final stage with the latent-action
head removed. Table~\ref{tab:hyperparams} lists the settings of both stages;
they differ only in batch size and step count, which follow from the corpus
sizes.

\begin{table}[h]
\centering
\caption{Optimization settings. Both stages train the full network with
identical optimizer, schedule and precision; the effective batch follows the
corpus size.}
\label{tab:hyperparams}
\small
\setlength{\tabcolsep}{6pt}
\begin{tabular}{lcc}
\toprule
Setting & Pretraining & Post-training \\
\midrule
Training clips     & 153{,}892 & 262 \\
Batch (per GPU $\times$ GPUs $\times$ accum.) & $64\times4\times3$ & $8\times4\times3$ \\
Effective batch    & 768 clips & 96 clips \\
Training steps     & 7{,}236 & 18{,}000 \\
Prefix             & $k = 3$, 9 generated & $k=1$, 11 generated \\
\midrule
Optimizer          & \multicolumn{2}{l}{AdamW, $\beta = (0.9, 0.999)$, weight decay $0.01$} \\
Learning rate      & \multicolumn{2}{l}{$1\times10^{-5}$, constant} \\
Gradient clipping  & \multicolumn{2}{l}{$1.0$, global norm} \\
Precision          & \multicolumn{2}{l}{bf16 mixed, FSDP with gradient checkpointing} \\
EMA decay          & \multicolumn{2}{l}{$0.99$, started at step 100} \\
Seed               & \multicolumn{2}{l}{$0$} \\
Clip geometry      & \multicolumn{2}{l}{45 pixel frames $\to$ 12 latent frames} \\
\bottomrule
\end{tabular}
\end{table}

\section{Additional Results}
\subsection{Benchmarking video VAEs for ultrasound video reconstruction}
All four autoencoders reconstruct ultrasound video accurately, spanning
$36.24$ to $36.62$\ dB PSNR and $0.900$ to $0.922$ SSIM, a spread of
$0.38$\ dB and $0.022$ respectively. Autoencoders trained on natural video
therefore transfer to ultrasound without retraining, despite speckle texture
and view-dependent acoustic artefacts having no counterpart in natural video.
Reconstruction quality does not discriminate between the four, and they
separate on cost instead. HunyuanVideo-1.5 requires $18.85$\,GB of peak memory
against $2.22$\,GB for Wan2.1 and runs at half the throughput, while giving
the worst LPIPS by a factor of $1.9$.

We adopt the Wan2.1-T2V-1.3B autoencoder, which has the lowest peak memory and
the highest throughput of the four while matching the best LPIPS at $0.033$
and trailing the best PSNR by $0.12$\ dB. Its SSIM is the lowest of the four,
by $0.022$, which does not affect the downstream evaluation: generation
metrics are computed against VAE-decoded ground truth rather than raw frames,
so reconstruction error cancels. The margin is ample in any case.
Reconstruction reaches $36.50$\ dB PSNR and $0.033$ LPIPS, against
$16.09$\ dB and $0.439$ for causal generation in Table~\ref{tab:tasks}, so the
autoencoder is not the bottleneck on generation quality.

\begin{table}[h]
\centering
\caption{Comparison of open-source video autoencoders for ultrasound video reconstruction.}
\label{tab:vae_comparison}
\small
\begin{tabular}{lccccc}
\toprule
Models & Time & Peak Memory & PSNR $\uparrow$ (dB) & SSIM $\uparrow$ & LPIPS $\downarrow$ \\
\midrule
Wan2.1-T2V-1.3B          & 0.71 clips/s & 2.22 GB  & 36.50 & 0.900 & 0.033 \\
Wan2.2-TI2V-5B-Diffusers & 0.68 clips/s & 3.71 GB  & 36.24 & 0.922 & 0.035 \\
HunyuanVideo-1.5         & 0.35 clips/s & 18.85 GB & 36.24 & 0.921 & 0.062 \\
LongCat-Video            & 0.68 clips/s & 2.36 GB  & 36.62 & 0.916 & 0.033 \\
\bottomrule
\end{tabular}
\end{table}

\subsection{Results on pretraining with latent actions}
\begin{figure}[h]
    \centering
    \includegraphics[width=\linewidth]{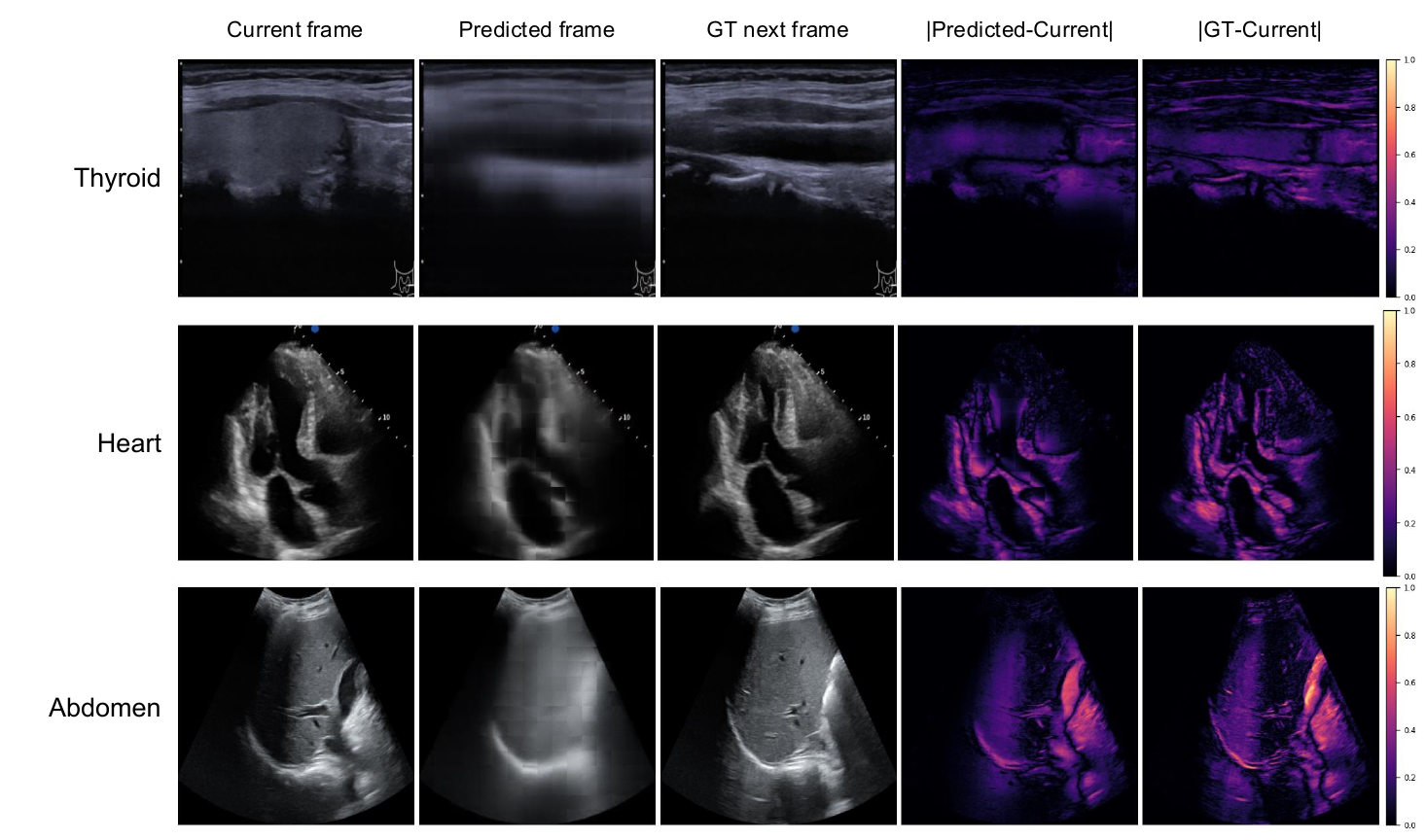}
    \caption{Latent action model predictions across organs. Rows correspond to
    thyroid, cardiac, and abdominal scans. Given only the current frame and the
    32-dimensional latent action, the decoder predicts the next frame; the
    rightmost two columns show the absolute difference of the predicted and of
    the ground-truth next frame from the current frame. }
    \label{fig:lam_multiorgan}
\end{figure}

\paragraph{Visualizations of next-frame prediction with latent actions}
Figure~\ref{fig:lam_multiorgan} visualizes latent action model predictions on
thyroid, cardiac and abdominal scans. The encoder infers a 32-dimensional
latent action from a frame pair, and the decoder reconstructs the next frame
from the current frame and that action alone, never seeing the next frame
itself, so any predicted change must have passed through the bottleneck. The
predicted frames recover the anatomy and its displacement but are smoother
than the ground truth, since speckle decorrelates between frames and is not
predictable from a 32-dimensional code. The two difference maps highlight the
same regions in all three organs, indicating that the latent action captures transition-relevant information beyond the appearance already provided by the source frame.

\begin{figure}[t]
\centering
\includegraphics[width=\linewidth]{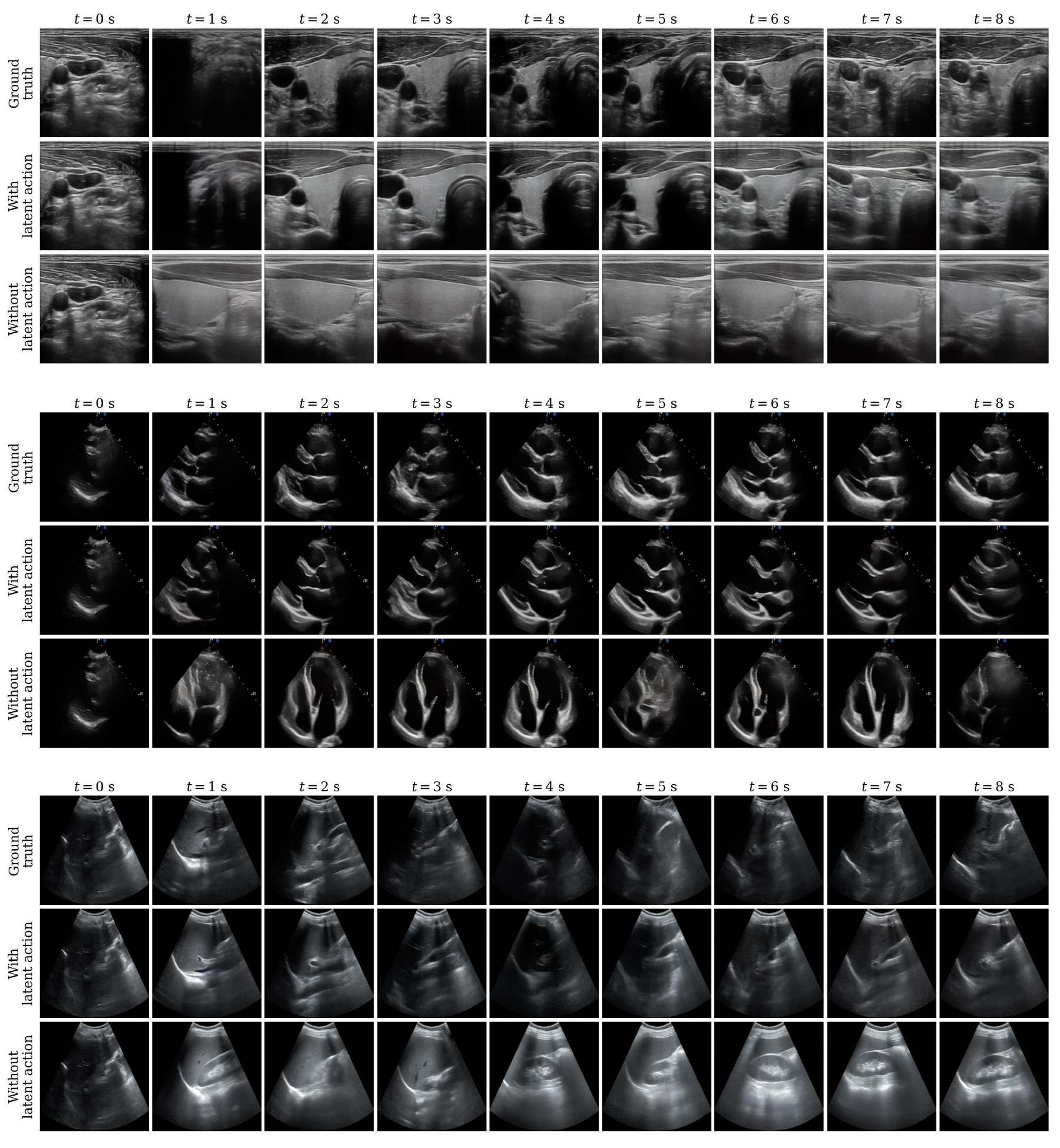}
\caption{Generation conditioned on latent actions. Rows from top to bottom are thyroid, cardiac, and abdominal scans. The model receives the first 9 ultrasound frames and the latent action sequence, then generates the remaining ultrasound frames. Generated frames follow the ground-truth anatomy as the probe moves across all three anatomies, showing that latent-action conditioning consistently captures useful transition dynamics across all three anatomical regions.}
\label{fig:gen_withlam}
\end{figure}

\paragraph{Results on video generations with latent actions.}
Table~\ref{tab:latent_action_organ} reports 8-second generation across the
three organs. Latent actions improve every metric on every organ, by
$4.74$\ dB PSNR, $0.114$ SSIM and $0.121$ LPIPS on average. The gain is
largest on abdominal scans ($+6.42$\ dB) and smallest on thyroid
($+3.21$\ dB), while the ordering of absolute difficulty is unchanged: thyroid
remains the hardest organ in both settings, reflecting its fine-scale
structure and its small share of the corpus. Since the action branch is
zero-initialized, the two models are identical at the start of training, so
the gap measures the information the action carries. The resulting performance gap quantifies the predictive value of latent transition information beyond visual context alone. Figure~\ref{fig:gen_withlam} shows the corresponding rollouts:
with latent actions, the generation tracks the ground-truth sequence, while
without them it drifts into a plausible but unrelated continuation.
\begin{table}[h]
\centering
\caption{Effect of latent actions on video generation across organs. We evaluate 8-second video
generation conditioned on 9 initial frames (3 latent frames) on abdominal, cardiac and thyroid scans.}
\label{tab:latent_action_organ}
\small
\setlength{\tabcolsep}{5pt}
\begin{tabular}{lccc|ccc}
\toprule
& \multicolumn{3}{c|}{w/o latent actions} & \multicolumn{3}{c}{w/ latent actions} \\
\cmidrule(lr){2-4} \cmidrule(lr){5-7}
Organ & PSNR $\uparrow$ & SSIM $\uparrow$ & LPIPS $\downarrow$
      & PSNR $\uparrow$ & SSIM $\uparrow$ & LPIPS $\downarrow$ \\
\midrule
Abdomen & 16.74 & 0.618 & 0.300 & \textbf{23.16} & \textbf{0.740} & \textbf{0.180} \\
Cardiac & 17.45 & 0.660 & 0.259 & \textbf{22.02} & \textbf{0.763} & \textbf{0.156} \\
Thyroid & 14.99 & 0.378 & 0.429 & \textbf{18.20} & \textbf{0.496} & \textbf{0.290} \\
\midrule
Average & 16.39 & 0.552 & 0.330 & \textbf{21.13} & \textbf{0.666} & \textbf{0.209} \\
\bottomrule
\end{tabular}
\end{table}

\subsection{Ablation studies: Effect of Latent-Action Pretraining on Pose-Conditioned Models}
Table~\ref{tab:lam_pretrain} examines whether latent-action pretraining benefits downstream
pose control. We compare two causal models with identical post-training and distillation,
differing only in whether latent actions were used during pretraining, and evaluate both
under five probe trajectories. Latent-action pretraining improves
every metric under every trajectory: PSNR increases by 0.40\,dB for the original pose,
0.42\,dB for Speed$\times$2, 0.46\,dB for the round trip and 0.42\,dB for 8-second rollouts,
with corresponding SSIM gains of 0.011--0.013 and LPIPS reductions of 0.005--0.008.
The gain is smallest under zero motion (0.21\,dB PSNR, 0.004 LPIPS), where the image remains
largely static and generation relies mainly on the conditioning frame. The benefit of
pretraining therefore grows when the probe moves, indicating that the dynamics learned from
unlabeled videos chiefly help the model render motion-induced appearance changes rather than
static anatomy. The round trip remains the most challenging setting for both models, yet the
pretrained model retains a 0.46\,dB advantage, the largest among all trajectories, suggesting
that pretraining also reduces error accumulation over extended bidirectional motion. Finally,
the advantage persists in 8-second rollouts that extend well beyond the 3-second training
horizon, showing that the improvement from pretraining is not limited to short clips.

\begin{table}[h]
\centering
\caption{Effect of Latent-Action Pretraining on Pose-Conditioned Models. Causal inference results under different probe trajectories.}
\label{tab:lam_pretrain}
\small
\begin{tabular}{llccc}
\toprule
Pretraining & Tasks & PSNR$\uparrow$ & SSIM$\uparrow$ & LPIPS$\downarrow$ \\
\midrule
\multirow{5}{*}{\shortstack{w/ latent\\actions}}
 & Original pose       & 16.09$\pm$0.95 & 0.386$\pm$0.054 & 0.439$\pm$0.044 \\
 & Zero motion         & 17.76$\pm$1.13 & 0.409$\pm$0.062 & 0.390$\pm$0.065 \\
 & Speed$\times$2      & 16.10$\pm$1.05 & 0.388$\pm$0.051 & 0.445$\pm$0.048 \\
 & Round trip          & 15.17$\pm$0.83 & 0.349$\pm$0.044 & 0.465$\pm$0.041 \\
 & Long (8\,s)         & 16.37$\pm$0.98 & 0.394$\pm$0.052 & 0.436$\pm$0.047 \\
\midrule
\multirow{5}{*}{\shortstack{w/o latent\\actions}}
 & Original pose       & 15.69$\pm$1.12 & 0.375$\pm$0.058 & 0.447$\pm$0.054 \\
 & Zero motion         & 17.55$\pm$1.01 & 0.398$\pm$0.055 & 0.394$\pm$0.058 \\
 & Speed$\times$2      & 15.68$\pm$1.06 & 0.375$\pm$0.052 & 0.453$\pm$0.052 \\
 & Round trip          & 14.71$\pm$0.91 & 0.337$\pm$0.050 & 0.470$\pm$0.050 \\
 & Long (8\,s)         & 15.95$\pm$1.07 & 0.382$\pm$0.057 & 0.444$\pm$0.050 \\
\bottomrule
\end{tabular}
\end{table}

\subsection{Results on causal inference}
\begin{figure}[t]
    \centering
    \includegraphics[width=\linewidth]{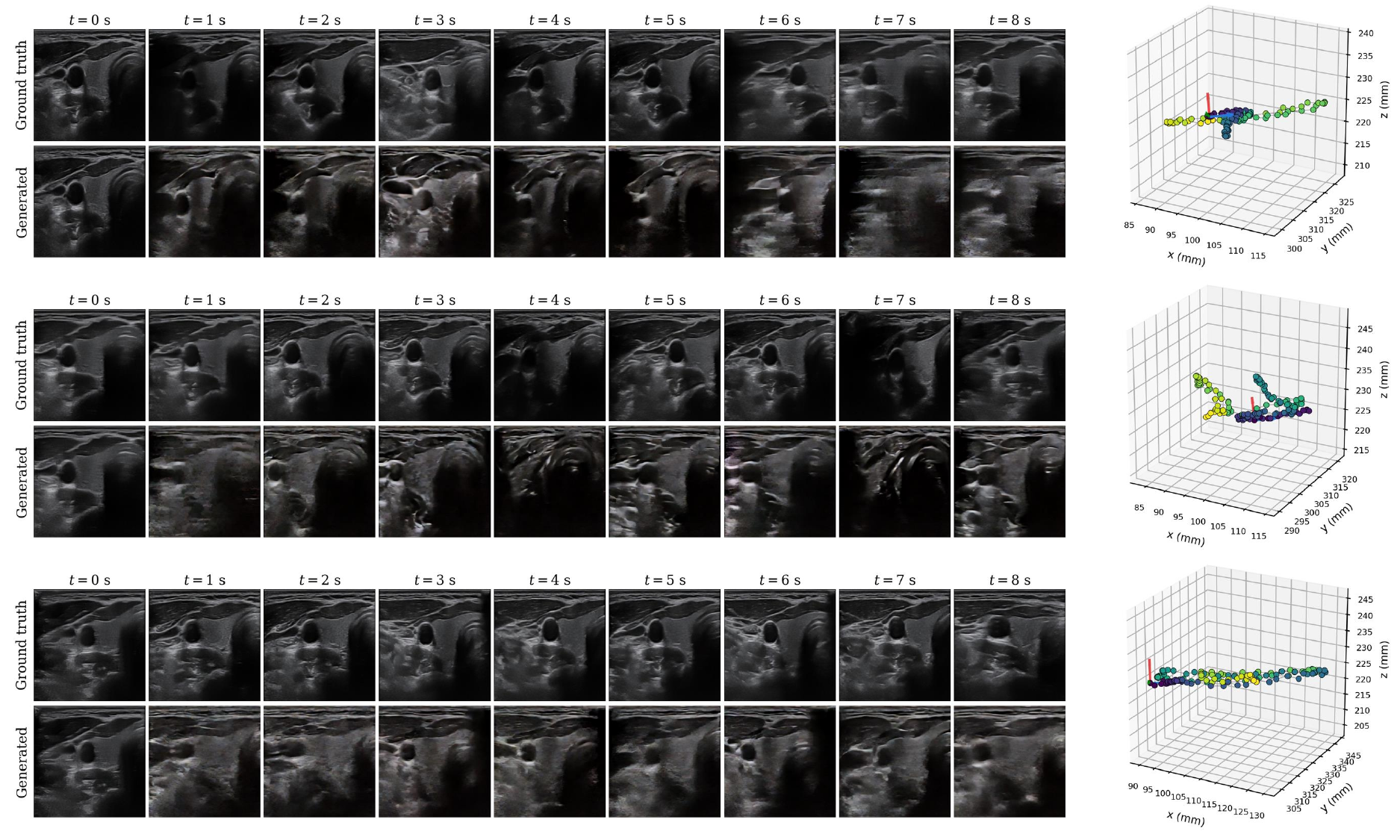}
    \caption{Eight-second rollouts from the causal model on three held-out scans. Each row is one scan: the model receives the initial frame and the measured 6-DoF pose sequence, then generates remaining frames autoregressively in four denoising steps. The right column shows the corresponding probe trajectory in the patient frame.}
    \label{fig:causal_rollout_traj}
\end{figure}

Figure~\ref{fig:causal_rollout_traj} shows 8-second rollouts on three held-out
scans. The causal student is trained on 3-second clips of 12 latent frames,
yet at inference it extends to 33 latent frames without retraining: the
generation still follows the ground-truth anatomy at $t = 8$\,s, well beyond
its training horizon.

\begin{figure}[h]
\centering
\includegraphics[width=0.7\linewidth]{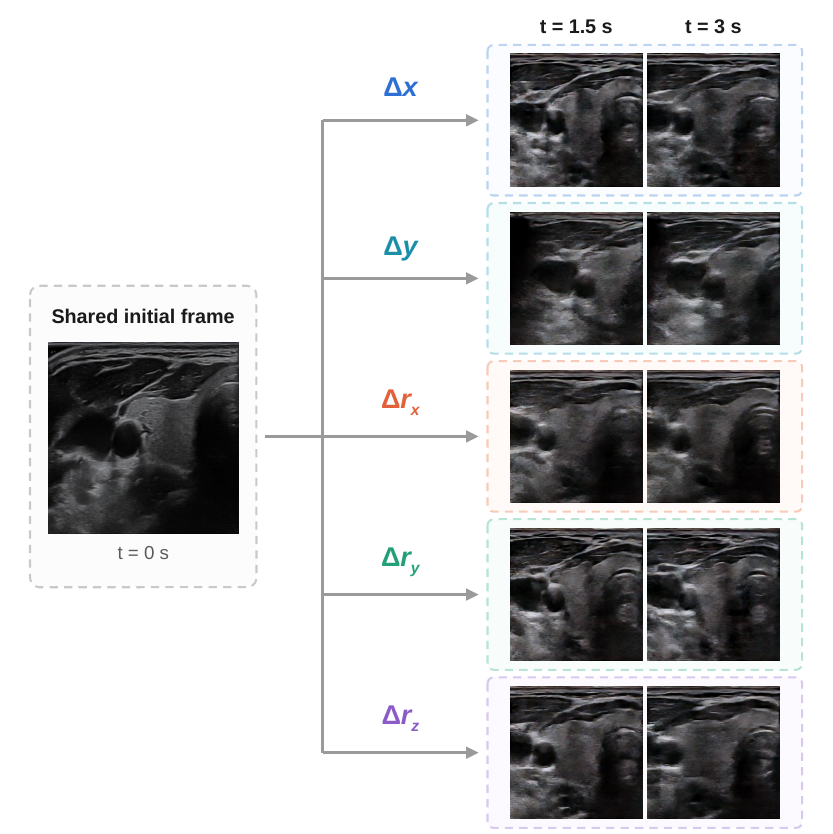}
\caption{Generation under customized actions. Starting from the same initial frame
($t=0$\,s), SonoWorld autoregressively generates 3-second rollouts, each
under a constant increment ($\Delta x$, $\Delta y$, $\Delta r_x$, $\Delta r_y$, or $\Delta r_z$) in a single degree of freedom.  
We set the constant increment to 1\,mm in translation and 1\textdegree{} in rotation.
The $x$, $y$, and $z$ axes correspond to the inferior--superior, left--right, and posterior--anterior
directions, respectively. }
\label{fig:gen_cust_pose}
\end{figure}

\section{User Study Interface}
\begin{figure}[h]
    \centering
    \includegraphics[width=\linewidth]{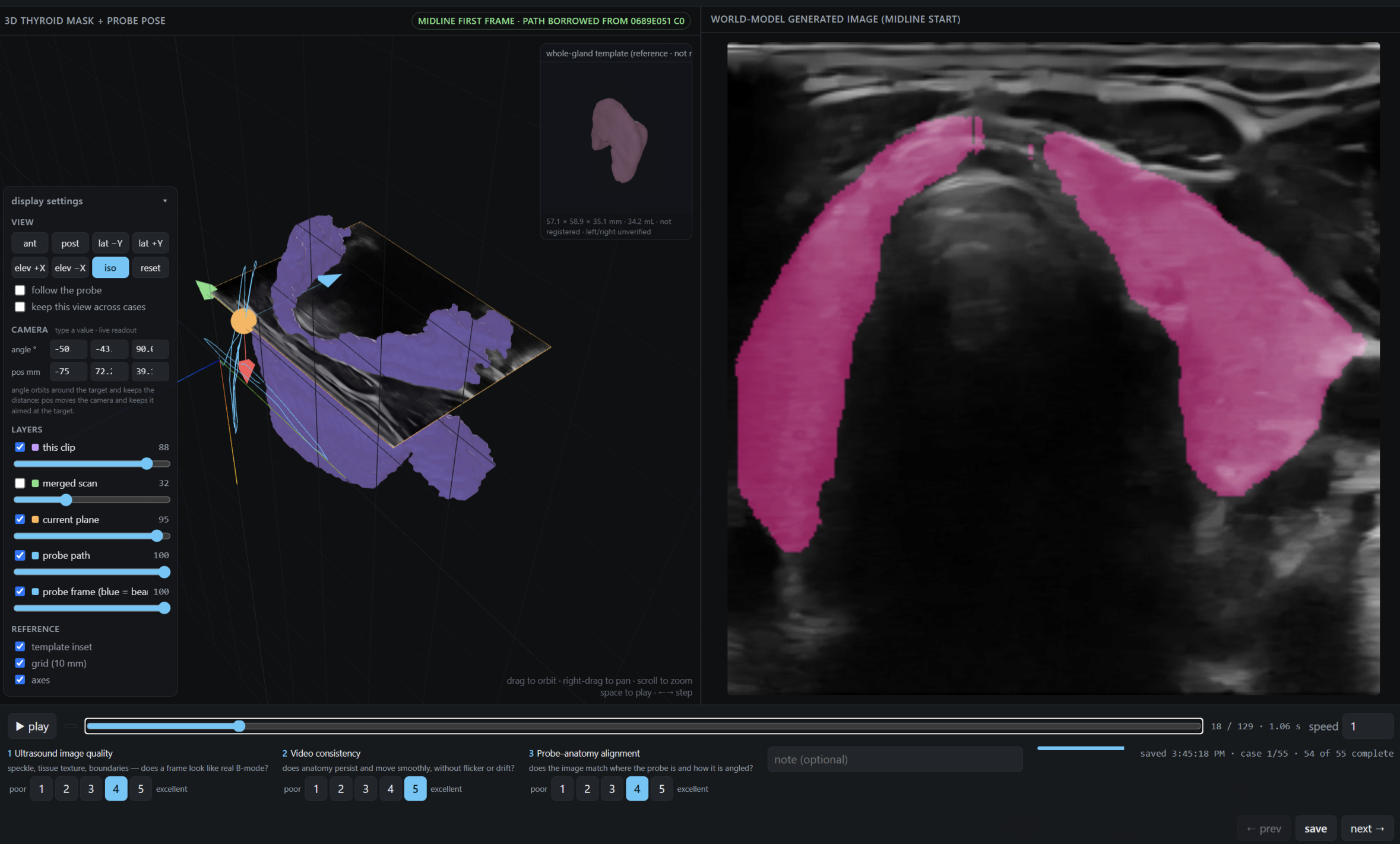}
    \caption{User study interface. Users are required to evaluate ultrasound video generated by actions according to three metrics and provide the evaluation score.}
    \label{fig:userstudygui}
\end{figure}

Figure~\ref{fig:userstudygui} shows the rating interface. The left panel
renders the segmented thyroid in 3D together with the probe path, the probe
frame and the current imaging plane, so that raters can see where each
generated frame is taken from, with a whole-gland template inset for
reference. The right panel plays the generated clip with its thyroid
segmentation overlaid. Raters scrub the timeline and score image quality,
video consistency and probe--anatomy alignment on 5-point scales.

\end{document}